\documentclass[10pt,letterpaper]{article}
\usepackage[T1]{fontenc}
\usepackage[letterpaper,left=0.72in,right=0.72in,top=0.72in,bottom=0.8in,footskip=27pt]{geometry}
\usepackage{amsmath,amssymb,newtxtext,newtxmath,microtype}
\usepackage[table]{xcolor}
\usepackage{graphicx,booktabs,xspace,enumitem}
\usepackage[numbers,sort&compress]{natbib}
\usepackage[font=small,labelfont=bf,labelsep=period,skip=5pt]{caption}
\usepackage{titlesec,flushend}
\usepackage[misc]{ifsym}
\usepackage{array}
\usepackage{tabularx}
\usepackage{multirow}
\usepackage{makecell}
\usepackage{pifont}
\usepackage{algorithm}
\usepackage{algpseudocode}
\definecolor{cvprblue}{rgb}{0.21,0.49,0.74}
\definecolor{clawteal}{HTML}{167D8D}
\definecolor{clawlight}{HTML}{EDF6F5}
\definecolor{draftgray}{HTML}{666666}
\definecolor{tablenavy}{HTML}{0071BC}
\colorlet{tablegroupbg}{tablenavy!5}

\definecolor{tableucsbg}{HTML}{E2ECEF}
\definecolor{tableucsink}{HTML}{00536B}
\definecolor{tableovobg}{HTML}{E1E8F3}
\definecolor{tableovoink}{HTML}{324779}
\colorlet{tablebeyondbg}{tableovobg}
\colorlet{tablebeyondink}{tableovoink}
\definecolor{tablevlmbg}{HTML}{E5E1F5}
\definecolor{tablevlmink}{HTML}{5A3477}
\newcommand{\tableband}[3]{\multicolumn{#1}{>{\columncolor{#2bg}}c}{\textcolor{#2ink}{\textit{#3}}}}
\newcommand{\tablemetric}[2]{\cellcolor{#1bg}\textcolor{#1ink}{#2}}
\newcommand{\resulttablerules}{\setlength{\heavyrulewidth}{0.6pt}\setlength{\lightrulewidth}{0.35pt}}
\newcommand{\tablegroup}[2]{\rowcolor{tablegroupbg}\multicolumn{#1}{l}{\textbf{\textit{#2}}}\\}
\newcommand{\method}{Dynamics-Claw\xspace}

\newcolumntype{Y}{>{\centering\arraybackslash}X}
\DeclareMathOperator{\Valid}{Valid}
\DeclareMathOperator{\Avail}{Avail}
\AtBeginDocument{\hfuzz=0pt\vfuzz=0pt}
\space
\definecolor{paperink}{HTML}{193F4B}
\definecolor{projectlabel}{HTML}{B62368}
\definecolor{projecturl}{HTML}{1864A5}
\usepackage[colorlinks,allcolors=paperink]{hyperref}
\hypersetup{pdftitle={Dynamics-Claw: Harnessing Streaming Video in the Physical 4D World},
  pdfauthor={Runyu Chen et al.}}
\titleformat{\section}{\large\bfseries}{\thesection.}{0.45em}{}
\titleformat{\subsection}{\normalsize\bfseries}{\thesubsection.}{0.4em}{}
\titlespacing*{\section}{0pt}{12pt plus 2pt minus 2pt}{5pt}
\titlespacing*{\subsection}{0pt}{8pt plus 2pt minus 1pt}{3pt}
\setlist[itemize]{leftmargin=1.35em,itemsep=2pt,topsep=3pt,parsep=0pt}
\newcommand{\aff}[1]{\textsuperscript{#1}}
\begin{document}
\twocolumn[{\begin{minipage}{\textwidth}
\centering
{\fontencoding{OT1}\fontfamily{ptm}\selectfont
\vspace*{55.4pt}
{\Large\bfseries Dynamics-Claw: Harnessing Streaming Video in the Physical 4D World\par}
\vspace{24pt}
{\large Runyu Chen\aff{1*}\quad Peiwei Lin\aff{1*}\quad Gaohui Liang\aff{1}\quad Li Luo\aff{2}\quad Liang Fan\aff{1}\quad Rongxin Gao\aff{1}\\
Xiaotong Tu\aff{1}\quad Yue Huang\aff{1}\quad Yuzhi Huang\aff{3}\quad Xinghao Ding\aff{1\Letter}\quad Kairun Wen\aff{4\textdagger}\par}
\vspace{6pt}
{\normalsize \aff{1}Xiamen University\quad \aff{2}University of Hong Kong\quad \aff{3}Tsinghua University\quad \aff{4}The Chinese University of Hong Kong\par}
\vspace{3pt}
{\small * Equal contribution\qquad \textdagger\ Project lead\qquad \Letter\ Corresponding author\par}
\vspace{3pt}
{\large\href{https://dynamics-claw.pages.dev/}{%
  \textcolor{projectlabel}{\textbf{Project Website:}}\enspace
  \textcolor{projecturl}{\texttt{https://dynamics-claw.pages.dev/}}}\par}}
\vspace{10pt}
\includegraphics[width=\linewidth]{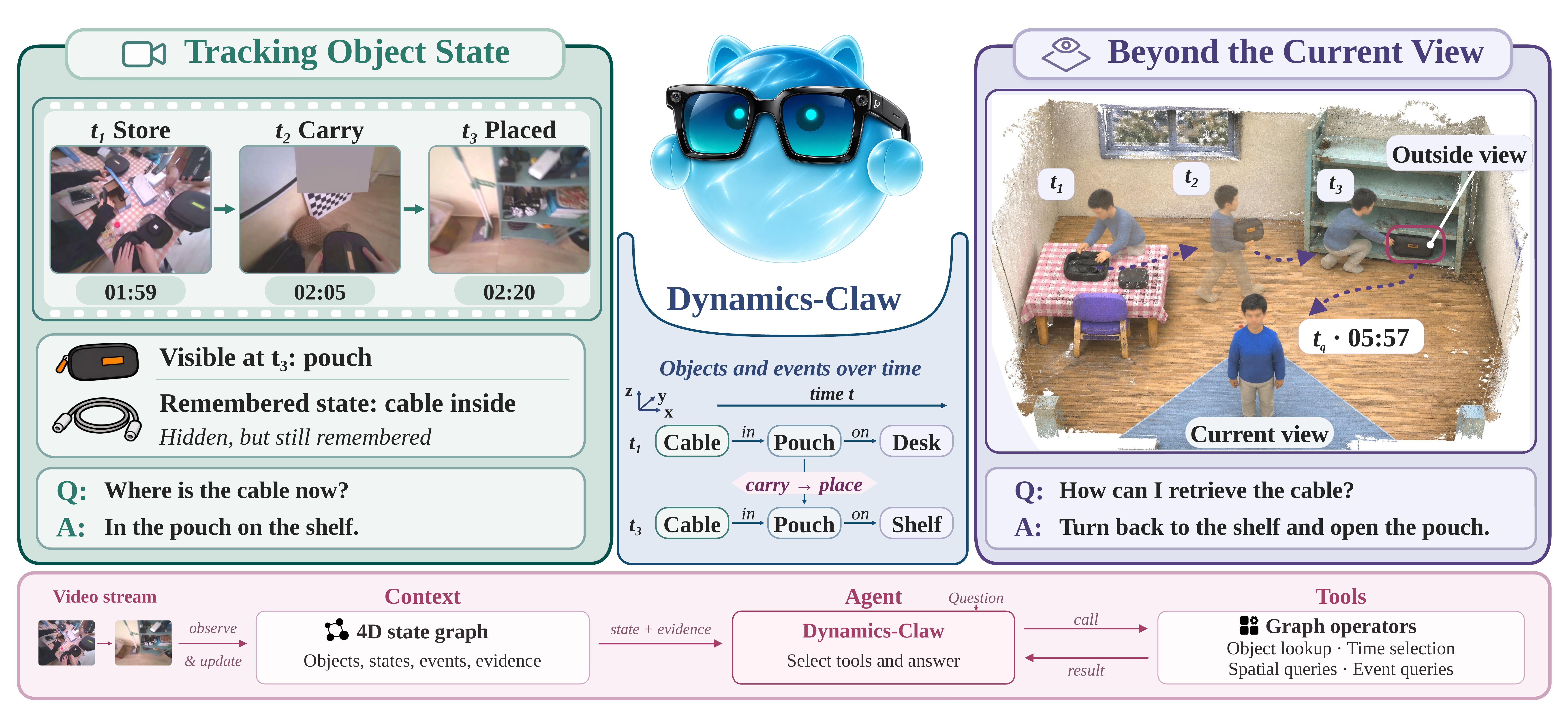}
\captionsetup{hypcap=false}
\captionof{figure}{\textbf{Overview of Dynamics-Claw, an agentic framework for streaming video understanding in the physical 4D world.} In this illustrative example, a cable remains inside a pouch as the pouch moves from a desk to a shelf. Streaming observations update a 4D state graph that serves as structured context. The agent calls graph operators for object lookup, temporal selection, spatial queries, and event queries to answer questions using maintained states and evidence.}\label{fig:teaser}
\par\vspace{7pt}\end{minipage}}]
\fontsize{10.6}{13.5}\selectfont
\setlength{\emergencystretch}{1.5em}
\setlength{\parskip}{0pt plus 1pt}
\renewenvironment{abstract}{%
  \fontencoding{OT1}\fontfamily{ptm}\normalsize\selectfont
  {\centering\large\bfseries Abstract\par}\vspace{12pt}%
  \itshape\noindent\ignorespaces}{\par}
\begin{abstract}
Understanding a physical scene from streaming egocentric video requires following how objects are handled, moved, and related to one another while the observer moves, often after the relevant objects have left view. Past observations may remain accurate records yet no longer describe the scene at the moment a question is asked. We introduce \textbf{\method}, an agentic framework for streaming video understanding in the physical 4D world. Its core is an event-and-object-aware 4D state graph, in which persistent object identities link time-scoped spatial states, interaction events, and supporting evidence; the graph serves as structured context for the agent. A harness coordinates two loops around a frozen vision-language model: the observation loop binds interaction events to tracked objects and updates their states as the stream arrives, while the query loop lets the agent call graph operators to retrieve, compute, and verify the states required by each question. We evaluate \method on UCS-Bench, OVO-S-Bench, and Beyond3D against streaming, spatial, and retrieval-based memories. It achieves the best overall accuracy on all three benchmarks, and ablations show that state-graph access, event processing, and graph-operator readout each contribute.
\end{abstract}
\space
\section{Introduction}
\label{sec:introduction}
Questions about a streaming video often concern what is no longer visible. In Fig.~\ref{fig:teaser}, a cable is placed in a pouch, and the pouch is later carried from a desk to a shelf. When asked where the cable is, neither the cable nor these actions are in view: the answer depends on earlier observations and on how the scene changed afterward. Streaming video understanding therefore requires memory that is built before the question arrives, under bounded observation and storage budgets~\cite{videollmonline,flashvstream,ucs}. The question is what this memory should retain.

Existing memories mainly keep records (Tab.~\ref{tab:memory-comparison}). Retrieval-based agents store descriptions, clips, or events for a model to interpret~\cite{m3agent,visualagenticmemory}, and token or cache memories preserve visual context~\cite{hermes,fluxmem}. Spatial maps and 4D scene graphs further record where and when objects were observed~\cite{ucs,daaam,graphitfirst}. In Fig.~\ref{fig:teaser}, records such as ``the cable is in the pouch'' and ``the pouch is on the desk'' are each correct, yet neither is the answer. The question requires \emph{state}: which facts about which object still hold at the time of the question. Foundation models for detection, tracking, geometry, and hand--object interaction (HOI) provide the identities, positions, and interactions from which such state can be built, but their outputs are scattered across frames and modules. Turning these outputs into state, and using that state to answer questions, both require decisions.

On the write side, the system must decide which changes are worth interpreting, which tracked object an interaction concerns, and which earlier state it invalidates. These decisions recur throughout the stream, before any question is known, and require coordinating perception, semantic interpretation by a vision-language model (VLM), and validated commits; we call this coordinating runtime a \emph{harness}. On the read side, different questions require different operations. When an object has been moved, the reader must select the state valid at the requested time; when the observer has turned, it must combine that state with the observer pose; when evidence is missing or ambiguous, it must inspect retained frames. The next operation depends on the question and on what the previous one returned: in Fig.~\ref{fig:teaser}, looking up the cable returns that it is inside the pouch, which redirects the query to the pouch's current position and then to the observer pose at the time of the question. Making these choices is the role of an \emph{agent}. The two are complementary: without maintained state, the agent's tools return evidence that must be reinterpreted at every query; without the agent, the state is a growing set of records that a VLM must read and compute over unaided.

\begin{figure*}[t]
\centering
\includegraphics[width=\textwidth]{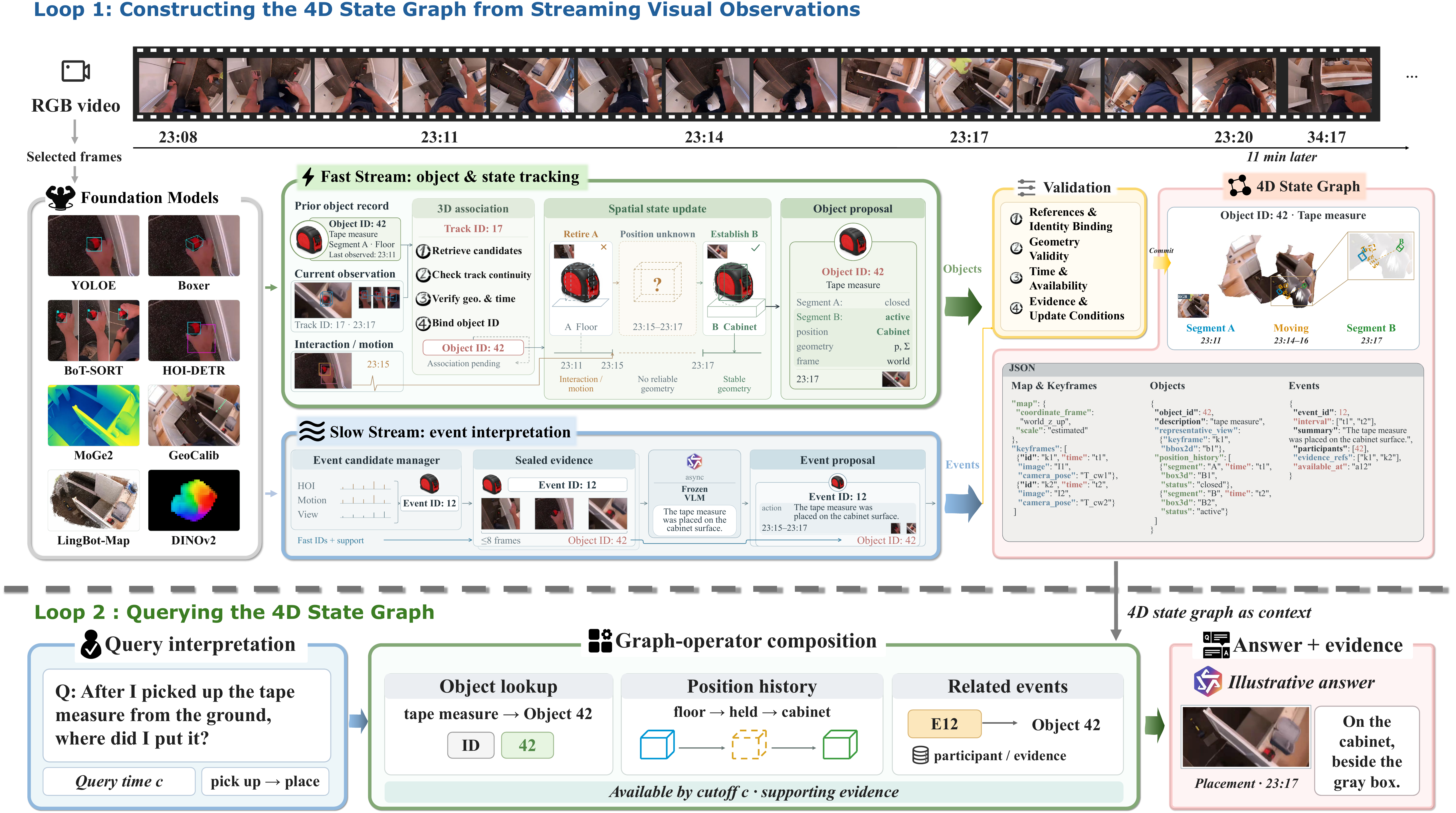}
\caption{\textbf{The overall architecture of Dynamics-Claw.} \textbf{Top:} Loop 1 constructs a 4D state graph from streaming visual observations: the fast stream maintains object identities and spatial states, while the slow stream interprets event evidence with a frozen VLM. Validated records connect persistent objects, position histories, events, camera poses, and retained frames, with temporal validity and information availability preserved. \textbf{Bottom:} Loop 2 uses an eligible graph snapshot as context and composes object lookup, position history, and event queries to answer questions with supporting evidence. The tape-measure visualization and JSON excerpt illustrate the graph representation.}
\label{fig:overview}
\end{figure*}

We introduce \method, an agentic framework for streaming video understanding in the physical 4D world. Its shared memory is an event-and-object-aware 4D state graph with three properties. First, persistent object identities are separated from position segments, so an interaction closes the current segment without deleting the object or its history. Second, events are bound to objects by matching HOI participants to tracked objects, so querying an object also recovers what happened to it. Third, every record carries a validity interval and evidence and is read through graph operators with defined semantics. Whereas text memories store descriptions for a model to read and 4D scene graphs timestamp observations, the state graph maintains states that can be updated and computed over. The observation loop builds and updates this graph as the stream arrives; the query loop supplies the relevant part of the graph as context and lets the agent call graph operators as tools (Fig.~\ref{fig:overview}).

\begin{table}[t]
\centering
\caption{\textbf{Comparison with existing video-memory methods.} Inputs are external modalities and poses. Object trajectories are explicit, queryable 3D motion paths; DAAAM and DirectMe follow their public implementations. HOI includes explicit hand--object descriptions; interaction updates apply interaction effects to maintained states.}
\label{tab:memory-comparison}
\begingroup
\normalsize
\resulttablerules
\setlength{\tabcolsep}{3pt}
\renewcommand{\arraystretch}{1}
\newlength{\memoryheadheight}
\settowidth{\memoryheadheight}{Interaction updates}
\newlength{\memorytrajectoryheadheight}
\settowidth{\memorytrajectoryheadheight}{Object trajectories}
\ifdim\memorytrajectoryheadheight>\memoryheadheight
\setlength{\memoryheadheight}{\memorytrajectoryheadheight}
\fi
\addtolength{\memoryheadheight}{10pt}
\newcommand{\memoryhead}[2]{\tablemetric{#1}{\rotatebox{90}{\makebox[\memoryheadheight][l]{\strut #2}}}}
\newcommand{\memoryyes}{\ding{51}}
\newcommand{\memoryno}{\ding{55}}
\newsavebox{\memorycomparisonbox}
\sbox{\memorycomparisonbox}{%
\begin{tabular}{lcc*{4}{w{c}{16pt}}cc}
\toprule
& \tableband{2}{tableucs}{Input and processing} & \tableband{4}{tableovo}{Scene modeling} & \tableband{2}{tablevlm}{Memory} \\
Method & \memoryhead{tableucs}{Input} & \memoryhead{tableucs}{Processing} & \memoryhead{tableovo}{Metric Geometry} & \memoryhead{tableovo}{Object trajectories} & \memoryhead{tableovo}{HOI} & \memoryhead{tableovo}{Interaction updates} & \memoryhead{tablevlm}{Representation} & \memoryhead{tablevlm}{Memory access} \\
\midrule
HERMES~\cite{hermes} & RGB & Streaming & \memoryno & \memoryno & \memoryno & \memoryno & KV cache & Cache attention \\
Video-RAG~\cite{videorag} & RGB + audio & Offline & \memoryno & \memoryno & \memoryno & \memoryno & Frames + extracted text & Text retrieval + frames \\
M3-Agent~\cite{m3agent} & RGB + audio & Streaming & \memoryno & \memoryno & \memoryyes & \memoryno & Entities + episodes & Iterative retrieval \\
AMEGO~\cite{amego} & RGB & Online build & \memoryno & \memoryno & \memoryyes & \memoryno & Tracklets + segments & Visual matching \\
DirectMe~\cite{ucs} & RGB & Offline/online & \memoryyes & \memoryno & \memoryno & \memoryno & Spatial graph & Spatial queries \\
DAAAM~\cite{daaam} & RGB-D + poses & Streaming & \memoryyes & \memoryno & \memoryno & \memoryno & 4D scene graph & Graph tools \\
EgoSG~\cite{graphitfirst} & RGB & Clip-wise build & \memoryno & \memoryno & \memoryyes & \memoryno & Temporal graph & Full-graph context \\
\midrule
\textbf{Dynamics-Claw} & RGB & Streaming & \memoryyes & \memoryyes & \memoryyes & \memoryyes & \textbf{4D state graph} & \textbf{Graph-operator composition} \\
\bottomrule
\end{tabular}%
}
\typeout{MEMORY-COMPARISON-NATURAL-WIDTH: \the\wd\memorycomparisonbox}
\typeout{MEMORY-COMPARISON-TARGET-WIDTH: \the\linewidth}
\resizebox{0.98\linewidth}{!}{\usebox{\memorycomparisonbox}}
\endgroup
\end{table}
\space

Our contributions can be summarized as follows:
\begin{itemize}
    \item We introduce \textbf{\method}, an agentic framework for streaming video understanding in the physical 4D world. It answers spatiotemporal questions about object states, interactions, and spatial relations in evolving scenes, including when the relevant objects are no longer visible.
    \item We design an event-and-object-aware \textbf{4D state graph} shared by two loops. The observation loop binds interaction events to persistent objects and updates their time-scoped spatial states; the query loop provides the graph as structured context and its operators as tools, so that the agent selects, computes, and verifies states instead of reconstructing them from retrieved history.
    \item We evaluate \method on UCS-Bench, OVO-S-Bench, and Beyond3D against streaming, spatial, and retrieval-based memories. \method achieves the best overall accuracy on all three benchmarks, and ablations show that state-graph access, event processing, and graph-operator readout each contribute to its performance.
\end{itemize}
\space
\section{Related Work}
\label{sec:related}

\subsection{Streaming Video Understanding}
Online video models process observations incrementally and retain a bounded context for later queries~\cite{videollmonline,flashvstream}. StreamForest uses event-oriented visual memory~\cite{streamforest}, HERMES manages a persistent KV cache~\cite{hermes}, and FluxMem compresses and consolidates visual tokens~\cite{fluxmem}. SimpleStream studies the strength of recent-frame evidence under a streaming protocol~\cite{simplestream}. These works motivate strong recency and compression baselines. Our focus is the semantics of memory updates and readout: whether explicit state maintenance helps when an earlier observation no longer directly determines the answer.

\subsection{Retrieval and Agent Memory}
M3-Agent and Visual Agentic Memory accumulate multimodal evidence and retrieve it to support reasoning~\cite{m3agent,visualagenticmemory}, and StreamingClaw provides a general framework for streaming interaction and memory~\cite{streamingclaw}. Such agents are typically training-free and attach an external memory to a frozen model, as \method does. The difference lies in what the memory holds and how it is accessed (Table~\ref{tab:memory-comparison}). Their memories store descriptions and episodes for the model to read, so the applicable state must be reconstructed at each query. \method maintains object states and their validity before a question arrives, and its agent reads them through graph operators with defined semantics. Reading the same graph as free-form text provides a control for this difference in our ablations.

\subsection{Object-Centric Spatial Memory}
AMEGO builds online interaction-centered object memories~\cite{amego}. Graph it first! constructs temporal scene graphs containing objects, attributes, relations, and interactions for egocentric question answering~\cite{graphitfirst}. DirectMe addresses spatial reasoning in continuous egocentric streams with a metric-semantic map~\cite{ucs}. Dynamic scene graphs, Hydra, and ConceptGraphs organize spatial structure for embodied reasoning~\cite{dsg,hydra,conceptgraphs}; 3D-Mem organizes scene observations for embodied inference~\cite{threedmem}. The scene-modeling dimensions in Table~\ref{tab:memory-comparison} situate our focus on persistent identities, separate position segments, and object-associated interaction history.

CATER and TCOW study tracking through containment and occlusion~\cite{cater,tcow}, while When Memory Lies examines outdated spatial memory~\cite{whenmemorylies}. Coordinate-frame composition~\cite{tf} and support dependencies~\cite{doyle} inform our spatial readout and correction semantics. \method combines these concerns in causal video memory: an interaction can retire a spatial estimate without deleting the object's identity, and later queries combine applicable states with time-stamped events and evidence.
\space
\section{Problem Formulation}
\label{sec:problem}
Frames $x_t$ arrive in temporal order. Query-independent ingestion maintains memory without access to future questions. A question specifies a target state time $t$ and an information cutoff $c$, and is answered by $a=R(M_{\leq c},q,t)$. The reader can access only records and retained evidence available by $c$, not future frames or discarded portions of the video. Methods allowed to reopen the full video prefix are reported separately.

A fact stores a predicate, its participants, a validity interval $[t_s,t_e)$, evidence references, a support set $S(f)$, and a status in $\{\textit{observed},\textit{inferred},\textit{unknown},\textit{conflict}\}$. Evidence records distinguish capture time from the time at which information became available to the system. Asking now about an earlier position permits $t<c$; asking what could have been known at that earlier time also restricts $c$. A late observation or identity binding can revise a later snapshot, but cannot improve an answer already issued at an earlier cutoff. Re-reading an old frame preserves its original timestamp.

The system separately bounds online working memory, per-query evidence, and VLM calls, while a persistent store may grow with the stream. Abstention counts as an error in multiple-choice accuracy. The task is therefore to maintain usable state under causal access and resource constraints, rather than to maximize the amount of stored history.
\space
\section{Method}
\label{sec:method}

\subsection{Overview}
\method organizes streaming video understanding into an observation loop and a query loop that share a 4D state graph (Fig.~\ref{fig:overview}). In the observation loop, the harness coordinates foundation models that provide object and geometry observations, a frozen VLM that interprets event evidence, and an executor that validates and commits structured records, all before future questions are known. In the query loop, the agent, driven by the same VLM, reads an eligible snapshot, chooses graph operators according to the question and their returned evidence, and formulates an answer.

\subsection{Event-and-Object-Aware 4D State Graph}
\label{sec:representation}
Answering questions about a changing scene requires state that can be located by object, updated locally when an interaction occurs, and composed across objects and time. The graph is organized around these operations. It instantiates 4D state memory through persistent object nodes, time-indexed spatial states, event records, and evidence links. Its central distinction is between \emph{which object} a record concerns and \emph{where that object was supported to be}. A persistent identity can connect several position segments as an object is moved. A segment contains geometrically supported observations over its interval; closing a segment preserves its history. Intermediate motion samples and uncertain positions remain distinguishable from a stable spatial estimate. The memory portion of Fig.~\ref{fig:overview} illustrates this separation for a tape measure.

Events retain their occurrence interval, semantic description, participant roles, and supporting frames. Evidence-backed participant links associate a local interaction with persistent objects. An unresolved participant can retain a local identity until a supported association becomes available. Thus, querying an object can recover both its position history and what happened to it, without replacing one by the other. Semantic relations, such as support or containment, are recorded when supported; their presence does not establish an exact relative 3D pose. Observer poses and object geometry carry their coordinate frames and uncertainty.

Records distinguish state time from information availability. At state time $t$ and cutoff $c$, a fact is usable only if
\begin{equation}
\begin{split}
\Valid(f;t,c)={}&[t_s(f)\leq t<t_e(f)]\\
&\land\Avail(f;c)\\
&\land\mathrm{Supported}(f;t,c).
\end{split}
\label{eq:valid}
\end{equation}
Here $\Avail$ checks committed-snapshot availability, and $\mathrm{Supported}$ checks the evidence and premises at the times required by the derivation. An event premise is evaluated at the event time; a spatial transform must apply at the requested state time. Unknown geometry or conflicting associations remain explicit rather than being replaced by an arbitrary state.

\subsection{Observation Loop and State Update}
\label{sec:harness}
The observation loop makes the write-side decisions: which changes to interpret, which object they concern, and which earlier state they invalidate. It combines continuous object maintenance with event-driven semantic updates. These processes share identities and evidence, but need not run at the same frequency.

\subsubsection*{Object Identity and Spatial State}
Per-frame detections and short tracks provide object candidates. Supported associations connect them to persistent IDs and bind geometry to the corresponding observations. New objects receive records; ambiguous associations remain unresolved. A short-track ID or a static-position segment is distinct from the persistent identity of the physical object.

The spatial frontend estimates observer pose and object geometry from the arriving RGB stream. Geometry retains its frame, time, and uncertainty. Interaction or reliable motion evidence retires an affected static position from current use, while retaining it as history. Subsequent stable observations establish a new position segment; continuity evidence determines whether it belongs to the same persistent object. A missing detection alone does not establish a new location. This separation allows identity and interaction history to persist through intervals in which a reliable current position is unavailable.

\subsubsection*{Event-to-State and Memory Update}
\label{sec:event}
\begin{figure}[t]
\centering
\includegraphics[width=\linewidth]{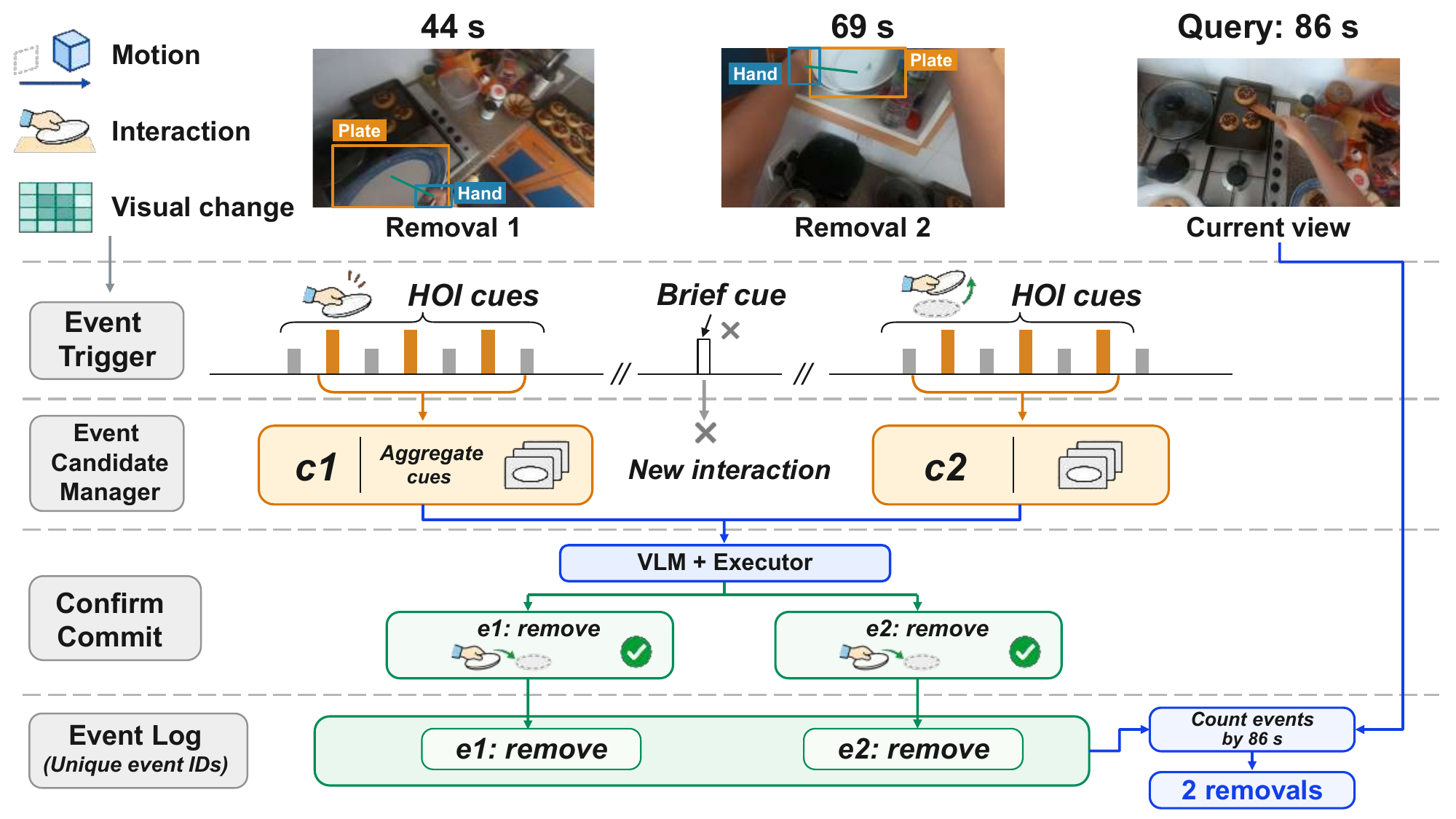}
\caption{\textbf{Event triggering and candidate management.} Motion, interaction, and visual-change cues support candidate formation, semantic interpretation, and memory updates. Two annotated UCS-Bench removals (39--45 and 65--71\,s) illustrate distinct event IDs and a count of two by 86\,s~\cite{ucs}. Photographs are video frames; boxes, cues, and event records illustrate the mechanism rather than measured system outputs.}
\label{fig:event-state}
\end{figure}

Event triggers identify changes worth interpreting from motion, hand--object interaction (HOI), and visual-change cues. An HOI signal indicates contact or interaction risk, while the action and displacement require their own evidence. The Event Candidate Manager groups repeated interaction cues, retains their temporal support and representative frames, and determines when a candidate is ready for semantic interpretation. Observer visual changes provide additional context. Figure~\ref{fig:event-state} illustrates why repeated cues from one interaction should be aggregated while two separate removals retain distinct event identities.

The frozen VLM interprets a candidate's selected frames and describes the event. Participant identity comes from programmatic association: matching the HOI object to a same-frame tracked object links the event to its persistent identity. The description, time interval, and evidence then become queryable as that object's interaction history. Ambiguous participants remain local, and later binding evidence is recorded with its own availability time. A language description alone does not establish a persistent identity.

Event-to-State connects these object-associated events with maintained state. Interaction and motion evidence can invalidate the old static-position segment; later reliable geometry supports the next one. Supported attributes and relations retain their temporal scope and evidence. The executor checks references, time, evidence, and applicable update conditions before committing records. Repeated evidence for the same episode does not create additional events or repeatedly extend an effect. When only endpoint states are observed, the record preserves that interval without inventing the unseen action.

\label{sec:revision}
New observations can correct a current estimate while preserving its supported history. Current conclusions depending on superseded facts must be recomputed or marked uncertain. An observed change does not by itself reveal when an unseen transition occurred. Queries read committed snapshots; a late event description or identity binding cannot enter an answer whose information cutoff precedes its availability.

\begin{figure*}[t]
\centering
\includegraphics[width=\textwidth]{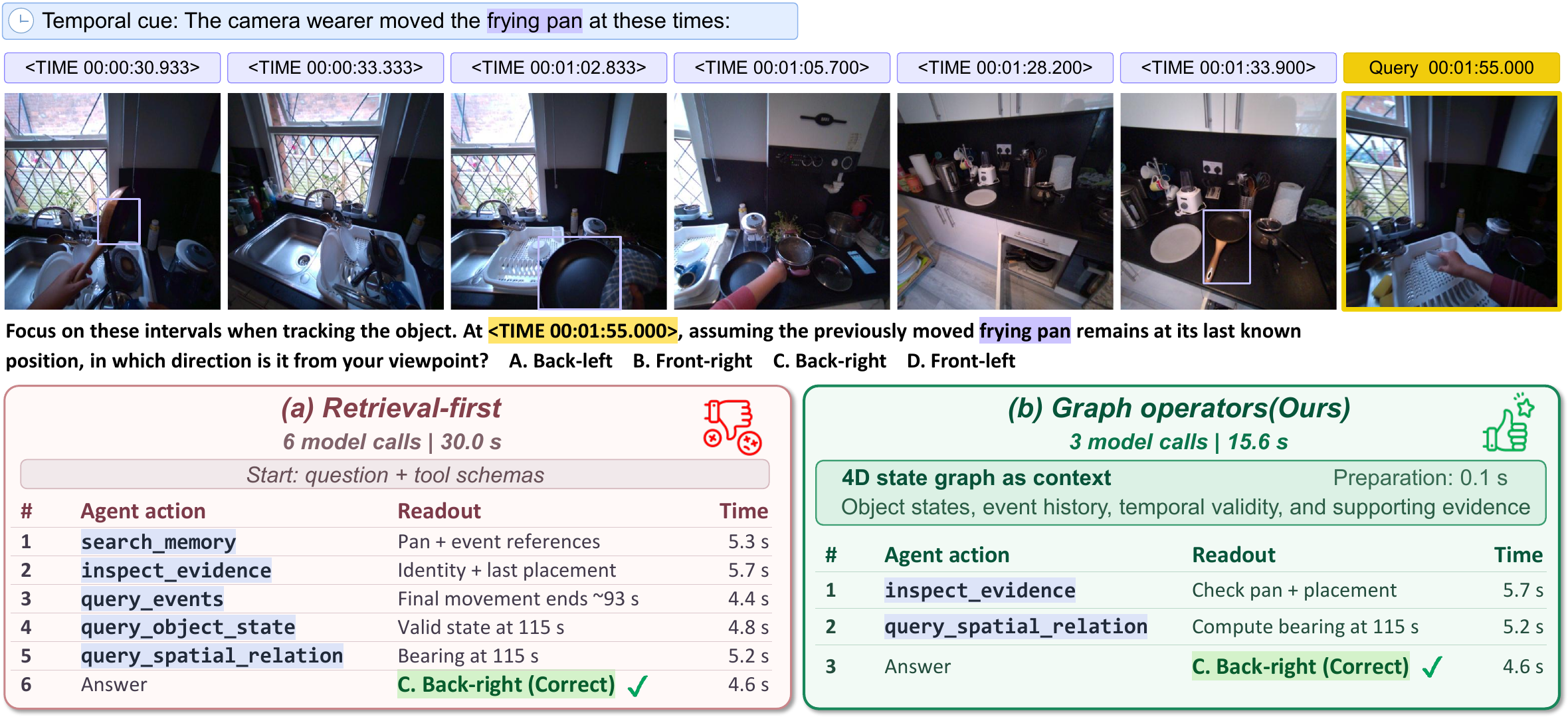}
\caption{\textbf{Graph context for tool-based spatial readout.} Video observations from HD-EPIC~\cite{hdepic} ground a query about the frying pan's direction at 115\,s, assuming it remains at its last placement. The retrieval-first path searches for relevant evidence, events, and object state. \method supplies a bounded 4D state-graph context containing object states, event history, temporal validity, and evidence links to guide evidence inspection and spatial-relation queries.}
\label{fig:memory-readout}
\end{figure*}

\subsection{Query Loop and Memory Readout}
\label{sec:composition}
The query loop turns a question into a sequence of memory operations. It identifies the requested objects or events, state time $t$, and information cutoff $c$, then selects the eligible committed snapshot. Before any tool call, the agent receives a bounded graph context from this snapshot, containing object states, event history, temporal validity, and evidence links. The context tells the agent what is already known, so that tool calls are spent on computation and verification rather than on locating the relevant objects and events. The VLM chooses operations according to the question and tool outputs: object lookup can lead to a position-history read, a participant link can lead to related events, and missing semantic detail can prompt inspection of a retained frame. This loop continues until the evidence supports an answer or the query budget is exhausted.

Graph operators are exposed to the agent as tools. The toolset includes memory search, which locates candidate objects and events; object-state query, which returns the state valid at time $t$; event query, which returns an object's interaction history with unique event identities; spatial-relation query, which computes direction or distance relative to the observer or another object; and evidence inspection, which opens retained frames for visual verification. Every output carries record IDs, temporal scope, and evidence references, which are returned with the answer as its supporting evidence.

Graph-operator composition is the sequence of tool calls the agent issues for a question. For the direction query in Fig.~\ref{fig:memory-readout}, the graph context already identifies the pan and its last placement, so the agent composes evidence inspection with a spatial-relation query; without the context, it must first search memory and query events and object state to recover the same object and state. Spatial relations combine the object's world-frame state with the observer pose at $t$, so observer motion changes the answer without changing the stored state. When geometry is missing, the tool returns an unknown metric result, while supported semantic relations such as containment remain usable. The VLM formulates the answer from tool outputs and inspected images. Query operations leave ingestion memory unchanged; later events, bindings, and observations cannot leak into an earlier information cutoff, so answers about past moments remain causal.

\begin{table*}[t]
\centering
\caption{\textbf{Spatial benchmark results (accuracy, \%).} DirectMe, HERMES, FluxMem, SimpleStream, Video-RAG, and Dynamics-Claw use Qwen3-VL-8B. $\dagger$: query-time prefix access. Configurations are in the supplement. The highest and second-highest scores in each column are highlighted in \colorbox[HTML]{FFB366}{dark orange} and \colorbox[HTML]{FFE5CC}{light orange}, respectively.}
\label{tab:main}
\footnotesize
\resulttablerules
\newcommand{\spatialbest}[1]{\cellcolor[HTML]{FFB366}{\textbf{#1}}}
\newcommand{\spatialsecond}[1]{\cellcolor[HTML]{FFE5CC}{#1}}
\setlength{\tabcolsep}{2pt}
\begin{tabularx}{\textwidth}{>{\raggedleft\arraybackslash}p{0.18\textwidth}@{\hspace{12pt}}p{0.11\textwidth}*{4}{Y}c*{4}{Y}c}
\toprule
 & & \tableband{5}{tableucs}{UCS-Bench} & \tableband{5}{tableovo}{OVO-S-Bench} \\
Methods & Memory & \tablemetric{tableucs}{P\&O} & \tablemetric{tableucs}{T\&M} & \tablemetric{tableucs}{P\&R} & \tablemetric{tableucs}{C\&Q} & \tablemetric{tableucs}{\textbf{Overall}} & \tablemetric{tableovo}{L1} & \tablemetric{tableovo}{L2} & \tablemetric{tableovo}{L3} & \tablemetric{tableovo}{L4} & \tablemetric{tableovo}{\textbf{Overall}} \\
\midrule
\tablegroup{12}{General-purpose VLMs}
Qwen3-VL-8B~\cite{qwen3vl} & frames & 40.41 & \spatialsecond{47.06} & \spatialsecond{49.83} & 42.04 & 44.11 & 43.40 & \spatialsecond{53.89} & \spatialbest{54.12} & \spatialbest{41.24} & \spatialsecond{48.08} \\
Qwen3.5-4B~\cite{qwen35} & frames & 24.08 & 41.12 & 44.22 & 23.19 & 31.07 & 40.22 & 51.48 & 45.88 & 39.42 & 44.54 \\
\tablegroup{12}{Spatial models}
Cambrian-S~\cite{cambrians} $\dagger$ & prefix & 35.56 & 39.41 & 46.61 & 36.12 & 38.58 & 38.79 & 48.70 & 44.44 & 31.39 & 41.64 \\
VST-RL~\cite{vst} $\dagger$ & prefix & 27.62 & 36.78 & 46.61 & 23.47 & 31.56 & 41.34 & 49.81 & 46.95 & \spatialsecond{40.88} & 44.83 \\
DirectMe~\cite{ucs} & metric map & 33.91 & 35.58 & 47.71 & 33.30 & 36.54 & 32.91 & 44.26 & 47.67 & 28.47 & 38.15 \\
\tablegroup{12}{Streaming memory}
Flash-VStream~\cite{flashvstream} & tokens & 29.55 & 34.78 & 45.31 & 31.29 & 34.13 & 35.14 & 40.56 & 39.43 & 36.50 & 37.75 \\
StreamForest~\cite{streamforest} & event tokens & 40.46 & 43.29 & 47.91 & 41.12 & 42.64 & 41.49 & 48.89 & 50.90 & 39.78 & 45.06 \\
HERMES~\cite{hermes} & KV & \spatialsecond{44.09} & \spatialbest{50.94} & 49.76 & \spatialsecond{43.41} & \spatialsecond{46.35} & 41.34 & 50.00 & 48.75 & 35.77 & 44.37 \\
FluxMem~\cite{fluxmem} & tokens & 39.63 & 46.77 & 49.14 & 41.44 & 43.52 & 45.15 & 52.59 & \spatialsecond{53.05} & \spatialsecond{40.88} & \spatialsecond{48.08} \\
SimpleStream~\cite{simplestream} & recent window & 39.68 & 40.55 & \spatialbest{51.61} & 34.25 & 40.12 & \spatialsecond{55.96} & 38.52 & 52.33 & 33.21 & 46.28 \\
\tablegroup{12}{Retrieval memory}
Video-RAG~\cite{videorag} & retrieval & 38.86 & 45.35 & 48.19 & 42.00 & 43.04 & 42.29 & 50.93 & 50.90 & 39.42 & 45.93 \\
\midrule
\textbf{Dynamics-Claw} & state graph & \spatialbest{51.70} & 42.26 & 42.98 & \spatialbest{55.00} & \spatialbest{49.25} & \spatialbest{57.23} & \spatialbest{64.26} & 35.48 & 39.78 & \spatialbest{53.14} \\
\bottomrule
\end{tabularx}
\end{table*}
\space

\subsection{Implementation Details}
The implementation connects YOLOE detection, BoT-SORT tracking~\cite{botsort}, HOI-DETR interaction cues, LingBot-Map and MoGe2 geometry, GeoCalib camera calibration, and Boxer object geometry to an executor and a logical memory store; DINOv2 features are reused from Boxer's visual encoder. Persistent object identity is maintained separately from native tracks and static-position segments. Event interpretation uses sealed evidence, and memory writes are coordinated by the main process. All models are pretrained and frozen, so \method requires no additional training. The VLM is Qwen3-VL-8B~\cite{qwen3vl} for Table~\ref{tab:main} and Qwen3.5-4B~\cite{qwen35} for Beyond3D and the ablations.

Memory is maintained incrementally. \method is not a mature real-time streaming system, but it runs with practical efficiency. Evicted images are marked inaccessible, so a stored pointer does not restore visual evidence. Interfaces, sampling rates, trigger thresholds, evidence budgets, and detailed time costs are given in the supplement.
\space
\section{Experiments}
\label{sec:experiments}
The evaluation compares methods on spatial and state reasoning, then isolates the contribution of each component through ablations.

\subsection{Experimental Setup}
UCS-Bench~\cite{ucs} and OVO-S-Bench~\cite{ovosbench} form the spatial axis. Beyond3D~\cite{beyond3d}, built on HD-EPIC~\cite{hdepic}, supplies the state axis through queries about relocated objects after they leave view, with visible controls for visibility checking. The evaluation covers the complete released benchmark: all 9,000 questions across eight types. The full-release out-of-sight records have a median query-to-last-visible separation of 18.5\,s (IQR 7--38\,s).

For \method and its ablations, each video is ingested once without access to questions or annotations, and each answer uses the snapshot available through its query timestamp, including the query frame. Ground-truth visibility, movement, and geometry are reserved for scoring.

Main comparisons retain each method's input-access assumptions; prefix readers may access the video through the query frame. Methods without their own models use Qwen3-VL-8B on UCS-Bench and OVO-S-Bench, and Qwen3.5-4B on Beyond3D. Core ablations use fixed development subsets of UCS-Bench and Beyond3D with Qwen3.5-4B, detailed in the supplement; Full and all variants share the same subsets, inputs, scoring, cutoffs, and budgets.

Tables report accuracy, with abstentions counted as incorrect. Beyond3D Macro averages all eight type accuracies; Visual reports visibility checking and Scene nearest-fixture accuracy. The supplement provides benchmark protocols, per-type scores, subset manifests, and detailed time costs.

\subsection{Main Results}
Table~\ref{tab:main} compares general-purpose VLMs, spatial models, streaming memories, retrieval-based methods, and \method on the complete UCS-Bench and OVO-S-Bench. \method achieves the best Overall accuracy on both benchmarks, 2.90 points above HERMES on UCS-Bench and 5.06 points above the strongest baselines on OVO-S-Bench. The gains concentrate on P\&O and C\&Q, by 7.61 and 11.59 points over the next best method, and on L1 and L2, by 1.27 and 10.37 points. On T\&M, P\&R, L3, and L4, methods that retain dense visual context, namely the base VLM, HERMES, and SimpleStream, remain stronger.

Table~\ref{tab:controlled} reports results on the complete Beyond3D benchmark, where general-purpose VLMs read the video prefix through the query time. \method achieves the best Temporal, Scene, 3D Spatial, and Macro accuracy. Its largest margins are on Temporal and 3D Spatial questions, which require knowing when an object last changed and where it lies relative to the observer or other objects. DirectMe and the composition baseline remain slightly higher on visibility checking.

The composition baseline ($\ddagger$) combines a spatial map with event-log retrieval and VLM reasoning. It uses the same identity and perception inputs as \method but leaves state reconstruction to the VLM at query time. Although it is comparable on visibility checking, its Temporal, Scene, and 3D Spatial accuracies stay at the level of the other baselines, and its Macro accuracy is 9.06 points lower than that of \method. Simply composing a spatial map with retrieved event logs therefore does not recover the state these questions require; the gain of \method comes from maintaining that state before the question arrives.

\ifdefined\resultstablesqueued\else
\providecommand{\beyondtablefloat}{table}
\begin{\beyondtablefloat}[t]
\centering
\caption{\textbf{Beyond3D results.} Accuracy (\%) on the full benchmark. Methods without their own models, including Dynamics-Claw, use Qwen3.5-4B; $\dagger$ follows Table~\ref{tab:main}; $\ddagger$: spatial map + event-log retrieval. The highest and second-highest scores in each column are highlighted in \colorbox[HTML]{FFB366}{dark orange} and \colorbox[HTML]{FFE5CC}{light orange}, respectively. Vis., Temp., and 3D denote Visual, Temporal, and 3D Spatial; Macro averages all eight types.}
\label{tab:controlled}
\footnotesize
\newcommand{\beyondbest}[1]{\cellcolor[HTML]{FFB366}{#1}}
\newcommand{\beyondsecond}[1]{\cellcolor[HTML]{FFE5CC}{#1}}
\setlength{\tabcolsep}{3.2pt}
\renewcommand{\arraystretch}{1.2}
\begin{tabularx}{\linewidth}{l*{5}{Y}}
\Xhline{0.8pt}
\textbf{Method} & Vis. & Temp. & Scene & 3D & \textbf{Macro} \\
\hline
Qwen3-VL-8B~\cite{qwen3vl} & 59.05 & 22.20 & 47.90 & 32.88 & 35.36 \\
Qwen3.5-4B~\cite{qwen35} & 59.80 & \beyondsecond{23.20} & 44.00 & 34.78 & 36.16 \\
Cambrian-S~\cite{cambrians} $\dagger$ & 53.60 & 21.80 & 50.90 & 33.20 & 35.11 \\
VST-RL~\cite{vst} $\dagger$ & 57.70 & 20.10 & 47.60 & 32.68 & 34.53 \\
DirectMe~\cite{ucs} & \beyondsecond{68.70} & 16.70 & 37.80 & 33.85 & 34.41 \\
Flash-VStream~\cite{flashvstream} & 50.95 & 18.35 & 43.10 & 31.00 & 31.84 \\
StreamForest~\cite{streamforest} & 65.05 & 21.20 & \beyondsecond{54.30} & \beyondsecond{35.25} & \beyondsecond{37.84} \\
HERMES~\cite{hermes} & 63.05 & 21.80 & 44.70 & 32.38 & 35.11 \\
FluxMem~\cite{fluxmem} & 61.90 & 22.45 & 45.20 & 32.95 & 35.48 \\
SimpleStream~\cite{simplestream} & 65.85 & 18.55 & 25.90 & 32.20 & 32.21 \\
Video-RAG~\cite{videorag} & 57.70 & 17.60 & 40.70 & 32.30 & 32.85 \\
Map + event retrieval $\ddagger$ & \beyondbest{69.05} & 17.65 & 40.40 & 32.95 & 34.57 \\
\midrule
\textbf{Dynamics-Claw} & 68.15 & \beyondbest{\textbf{31.35}} & \beyondbest{\textbf{55.60}} & \beyondbest{\textbf{40.65}} & \beyondbest{\textbf{43.63}} \\
\Xhline{0.8pt}
\end{tabularx}
\end{\beyondtablefloat}
\space
\begin{table*}[t]
\centering
\caption{\textbf{Component ablations on the UCS-Bench and Beyond3D development subsets.} The highest and second-highest scores in each column are highlighted in \colorbox[HTML]{FFB366}{dark orange} and \colorbox[HTML]{FFE5CC}{light orange}, respectively. \textit{S}, \textit{E}, and \textit{G} denote state-graph access, event processing, and graph-operator readout; \textit{G} requires \textit{S}. The subsets, backbone, and query cutoffs are shared. Vis., Temp., and 3D denote Visual, Temporal, and 3D Spatial; Beyond3D Macro averages all eight types.}
\label{tab:ablations}
\newcommand{\ablationbest}[1]{\cellcolor[HTML]{FFB366}{\textbf{#1}}}
\newcommand{\ablationsecond}[1]{\cellcolor[HTML]{FFE5CC}{#1}}
\setlength{\tabcolsep}{4pt}
\renewcommand{\arraystretch}{1.2}
\resizebox{0.8\textwidth}{!}{%
\begin{tabularx}{\textwidth}{l*{10}{Y}}
\Xhline{1pt}
\multicolumn{1}{l}{\multirow{2}{*}[-0.55ex]{\textbf{Configuration}}} &
\multicolumn{5}{c}{\textbf{UCS-Bench}} &
\multicolumn{5}{c}{\textbf{Beyond3D}} \\
\cmidrule(lr){2-6}\cmidrule(lr){7-11}
 & P\&O & T\&M & P\&R & C\&Q & \textbf{Overall} & Vis. & Temp. & Scene & 3D & \textbf{Macro} \\
\hline
w/o \textit{S}, \textit{E}, \textit{G} & 38.26 & 42.86 & 48.63 & 38.38 & 41.16 & 62.00 & 20.00 & 44.50 & 33.00 & 34.81 \\
w/ \textit{E} only & 40.92 & 48.57 & 48.29 & 41.90 & 44.24 & 64.25 & 27.00 & 47.50 & 33.50 & 37.47 \\
w/ \textit{S} only & 42.62 & 44.29 & \ablationsecond{50.34} & 39.44 & 43.25 & 63.25 & 22.00 & 48.00 & 34.88 & 36.84 \\
w/ \textit{S} + \textit{E} & 44.79 & \ablationbest{51.43} & \ablationsecond{50.34} & \ablationsecond{43.84} & \ablationsecond{46.89} & \ablationbest{65.50} & \ablationsecond{29.50} & \ablationsecond{51.50} & 35.13 & \ablationsecond{39.56} \\
w/ \textit{S} + \textit{G} & \ablationsecond{47.70} & 45.71 & \ablationbest{52.74} & 41.37 & 45.96 & 63.75 & 22.75 & 48.00 & \ablationsecond{37.38} & 38.34 \\
\textbf{w/ \textit{S} + \textit{E} + \textit{G} (Full)} & \ablationbest{49.88} & \ablationsecond{50.57} & \ablationbest{52.74} & \ablationbest{47.18} & \ablationbest{49.60} & \ablationsecond{64.75} & \ablationbest{31.25} & \ablationbest{57.50} & \ablationbest{38.63} & \ablationbest{42.41} \\
\Xhline{1pt}
\end{tabularx}
}
\end{table*}
\space
\fi

\subsection{Ablation Study}
Table~\ref{tab:ablations} lists the six valid combinations of state-graph access (\textit{S}), event processing (\textit{E}), and graph-operator readout (\textit{G}). With \textit{E} enabled, the observation loop processes HOI and observer event candidates; with \textit{E} disabled, both event branches and their selected frames and captions are removed, while upstream perception, tracking, and basic object memory remain. At each setting of \textit{E}, the flat reader accesses retained keyframes and visual descriptions, the \textit{S} reader additionally reads graph records, and the \textit{S}+\textit{G} reader also calls graph operators. Readers within each setting share ingestion and query snapshots. Graph operators require a graph, so \textit{G}-only and \textit{E}+\textit{G} are excluded.

Each component contributes on both benchmarks. State-graph access improves UCS Overall by 2.09 and 2.65 points without and with event processing, and Beyond3D Macro by 2.03 and 2.09. Event processing yields the largest single-component gains, 3.08--3.64 points on UCS and 2.66--4.07 on Beyond3D, with the clearest improvements on T\&M and Temporal questions. Graph operators add 2.71 points on UCS in both settings and 1.50--2.85 on Beyond3D, mainly on P\&O and 3D Spatial, with small decreases on T\&M and visibility checking when events are enabled. The \textit{S} reader treats the graph as text for the VLM to read, as in memories that attach descriptions to a frozen model; the additional gain from \textit{G} reflects the value of operating on maintained states rather than reading them. The full configuration improves over the variant without all three components by 8.44 points on UCS and 7.60 on Beyond3D.

\subsection{Analysis and Efficiency}
\label{sec:analysis}
Fig.~\ref{fig:memory-readout} shows a recorded execution on an HD-EPIC query about the frying pan's direction at 115\,s. A retrieval-first agent, starting from the question and tool schemas, answers correctly with six model calls in 30.0\,s. With the 4D state graph as context, \method already has the pan's last placement and its validity, and answers correctly with three model calls in 15.6\,s. The saving comes from the memory search, event query, and object-state query that the context makes unnecessary (Sec.~\ref{sec:composition}), halving both the number of calls and the latency. Detailed ingestion and query time costs are reported in the supplement.
\space
\section{Limitations}
\label{sec:limitations}
\method depends on identity associations, interaction cues, and geometry estimates. A wrong participant binding can attach an event to the wrong object; a missed interaction can leave an outdated spatial estimate in current use. Early errors can persist without later evidence supporting correction. Events outside the observed stream and uncertain actions leave gaps in memory. Metric geometry is object-level and subject to drift. Retained imagery introduces storage and privacy costs, and replay-based question answering does not establish successful assistance during real interaction.

Out-of-view answers rely on persistence assumptions between supported observations. Long occlusions, rapid motion, and complex interactions can break identity continuity or invalidate those assumptions. Beyond3D evaluation targets the complete release with method-specific input access, while the supplemental HD-EPIC tasks use a video-end adaptation. We report runtime over the measured stream lengths. Incremental maintenance alone does not establish bounded persistent storage or real-time processing.
\space
\begingroup
\SetTracking{encoding=*,shape=*}{-12}\lsstyle
\section{Conclusion}
\label{sec:conclusion}
\method treats streaming video understanding as maintaining and querying the state of a physical scene. A harness turns perception and interaction cues into an event-and-object-aware 4D state graph before questions arrive, and an agent answers by calling graph operators over this state and the retained evidence. Results on three benchmarks and component ablations show that maintained state, event processing, and graph-operator readout each contribute to spatiotemporal question answering, particularly for objects that have changed or left view. In future work, we plan to improve real-time operation, strengthen identity association under long occlusions, and extend the framework to embodied settings such as robot egocentric streams, where the maintained state can also guide physical actions.
\space
\par\endgroup\pagebreak
\clearpage
\raggedbottom
\raggedend
\end{document}


\raggedbottom
\maketitlesupplementary
\renewcommand{\thesection}{\Alph{section}}
\renewcommand{\thetable}{S\arabic{table}}
\renewcommand{\thefigure}{S\arabic{figure}}
\begin{strip}
\centering
\includegraphics[width=\linewidth]{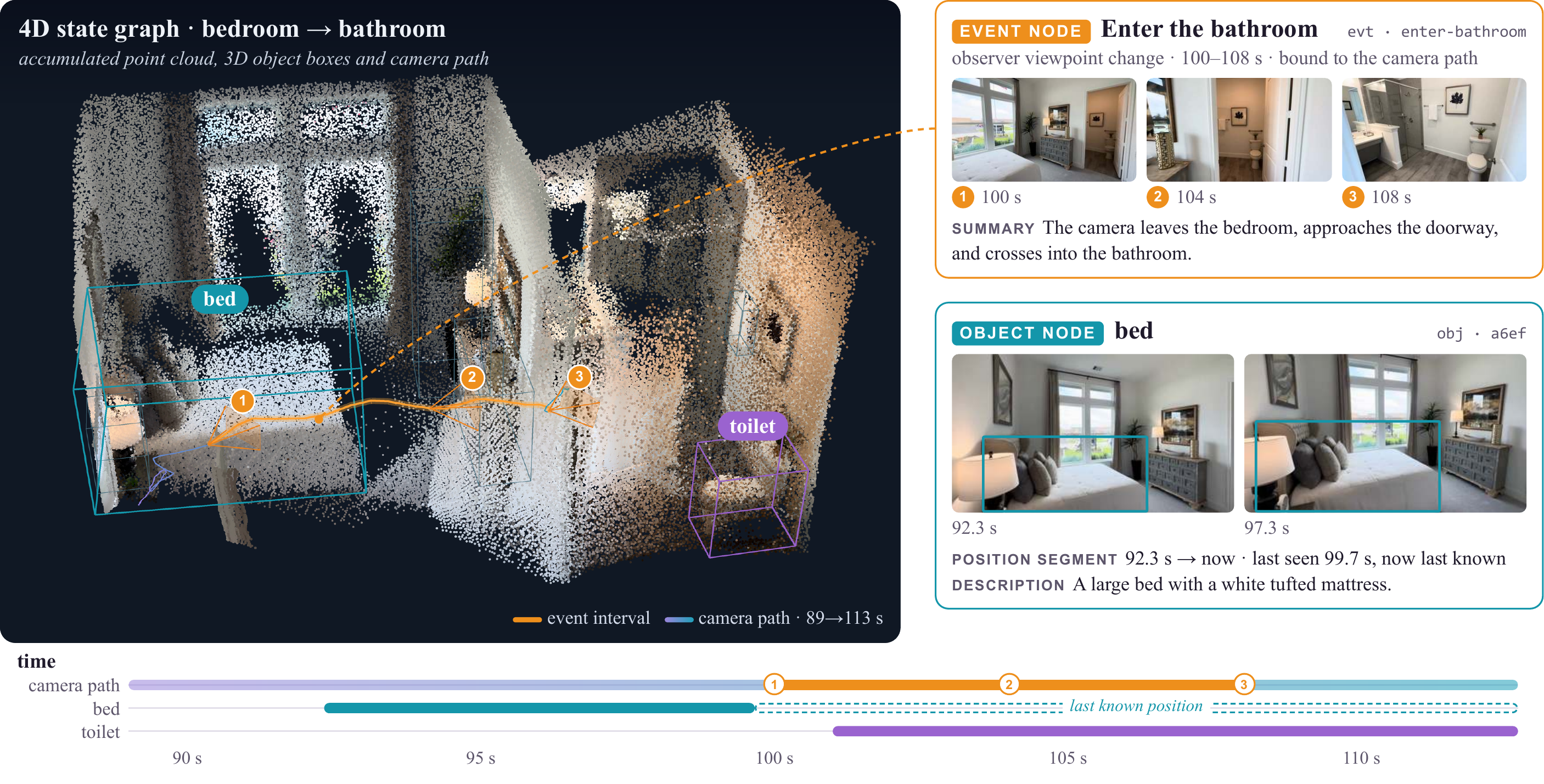}
\captionsetup{hypcap=false}
\captionof{figure}{\textbf{4D state graph of a walk from a bedroom into a bathroom.} Left: the accumulated point cloud with the 3D boxes of persistent objects and the camera path colored by time; the orange segment and camera frusta (1--3) mark the interval of the event node. Right: the event node \emph{Enter the bathroom} (100--108\,s) with its three representative frames, and the object node \emph{bed} with two representative views and its position segment. Bottom: the time span of each record; after its last observation at 99.7\,s, the bed is kept at its last known position. The dense point cloud is rendered for visualization; the memory stores object boxes, camera poses, events, and evidence frames.}
\captionsetup{hypcap=true}
\label{fig:supp-teaser}
\vspace{10pt}
\suppcontentsfirstpage
\end{strip}
\clearpage
\section{Method Details}
\label{sec:supp-method}

\subsection{Graph Records and Information Availability}
\label{sec:supp-records}

\paragraph{Records.}
All memory is stored as append-only records. Every record carries a record ID, a kind, an optional entity ID, an event time, an optional end of validity, an availability time, a commit time, a revision, an optional reference to the record it supersedes, its source records, and a typed payload. A record is never edited in place; a correction is a new record that supersedes the old one, so earlier versions remain readable. Table~\ref{tab:supp-records} lists the record types used by \method.

\begin{table*}[t]
\centering
\caption{\textbf{Record types in the 4D state graph.} Every record also carries the common fields described in Sec.~\ref{sec:supp-records}.}
\label{tab:supp-records}
\footnotesize
\renewcommand{\arraystretch}{1.15}
\begin{tabularx}{\textwidth}{lX}
\toprule
\textbf{Record} & \textbf{Main payload fields} \\
\midrule
Persistent object & Object ID, identity status, identity basis. \\
Track binding & Short-track ID, persistent object ID, position-segment ID. \\
Object box sample & 3D box (center, dimensions, rotation), detection confidence, reliability flag, position-segment ID, coordinate frame, scale status. \\
Object observation & 2D box, visibility and clarity of the observation. \\
Object alias & Merged object ID and canonical object ID. \\
Interaction event & Event type, summary, start and end times, participant roles, supporting frames, end reason. \\
Participant binding & Event participant, persistent object ID or unresolved, supporting frames, identity revision, identity availability time, previous binding. \\
Observer motion event & Start, last observed and end times, lifecycle status, dependency records. \\
Evidence & Retained frame or crop, capture time, checksum, retention status. \\
Semantic relation & Predicate (e.g., on, in), moved object, destination object, validity interval starting at the observed placement, supporting event and frames. \\
\bottomrule
\end{tabularx}
\end{table*}

\paragraph{Validity and status.}
Each fact has a status in \{\textit{observed}, \textit{inferred}, \textit{unknown}, \textit{conflict}\}. A fact is \textit{observed} when it is directly supported by evidence at the requested time and \textit{inferred} when it follows from other supported facts. When no record supports a state at the requested time and none can be inferred, the state is \textit{unknown}. Object positions are an exception: when an object is not currently observed, its last observed position is returned with its timestamp and marked as stale rather than replaced by \textit{unknown} (Sec.~\ref{sec:supp-update}). A static position is valid from the opening of its position segment until the segment is closed, and a semantic relation such as ``the cable is in the pouch'' until the object is seen being taken out; after that end, the position or relation is kept as history and is no longer used as the object's current location. At read time, a record is first checked for validity at the requested time and then for whether its supporting records are still valid. If two valid records place the same object in different locations, the result is \textit{conflict}.

\paragraph{Geometry.}
Object boxes and observer poses are expressed in one world frame per video, defined by the first camera and aligned with gravity (z up). LingBot-Map estimates camera intrinsics, camera poses, and relative depth. MoGe2 aligns the relative scale of LingBot-Map to metric scale with a factor fitted once per video on prefix frames, and GeoCalib estimates the gravity direction jointly over a five-frame prefix. Every geometric record carries its coordinate frame, scene, and scale revision. Uncertainty is represented by a status label (for example, reliable, stale, or scale estimated) and the detection confidence; covariances are not estimated.

\paragraph{Information availability and snapshots.}
Each record has three times: the event time, at which the observed state or interaction occurred; the availability time, at which the system could use the record; and the commit time, at which it entered memory. For an event interpreted by the VLM, the availability time follows the completion of interpretation, not the capture of its frames. Every commit increments a global revision. A query with information cutoff $c$ reads the snapshot containing the records whose revision does not exceed the current revision, whose availability, commit, and event times do not exceed $c$, and which are not superseded within that snapshot. A late description, identity binding, or geometric correction therefore enters only snapshots whose cutoff follows its availability time.

\paragraph{Evaluation timeline.}
For a question at time $T_q$, the system first processes all sampled frames up to $T_q$, freezes the memory at that point, and answers from the frozen snapshot; it then continues with later frames. When a video has several questions, its prefix is ingested only once. Background event interpretations and captions enter the snapshot only if they were committed before it was frozen; results still queued or being generated are used only by later questions.

\paragraph{Terminology.}
\emph{4D} refers to object-level 3D states indexed by time. The \emph{harness} is the runtime that coordinates perception, VLM interpretation, and validated commits. \emph{Causal} refers to information access: an answer at cutoff $c$ uses only information available by $c$. It does not denote physical cause-and-effect inference.

\subsection{State Update Rules}
\label{sec:supp-update}

Algorithm~\ref{alg:supp-write} summarizes one step of the observation loop. The main stream is processed at 5\,fps and the event branch samples it at 1\,fps.

\begin{algorithm}[t]
\small
\caption{Observation loop (write side)}
\label{alg:supp-write}
\begin{algorithmic}[1]
\Require Frame $x_t$ with event time $t$
\State Detect object boxes; detect hands and hand--object relations
\State Remove hand and forearm proposals; estimate camera pose and depth
\State Track proposals with BoT-SORT and lift them to 3D boxes
\State Associate tracks with persistent objects; create an object for an unmatched confirmed track
\State Close the position segment of an object with confirmed contact or reliable motion
\State Append stable box samples to the current position segment, or open a new one
\If{$t$ falls on the 1\,fps event tick}
    \State Update HOI candidates from hand--object cues
    \State Update observer candidates from visual-change and camera-motion cues
    \For{each sealed candidate with available evidence}
        \State Freeze before/key/after frames and participant bindings
        \State Submit the frames to the frozen VLM
    \EndFor
\EndIf
\For{each completed interpretation}
    \State Validate references, times, and evidence; commit the event and its binding
    \State Apply the event's supported effects to the bound objects' states
\EndFor
\end{algorithmic}
\end{algorithm}

\paragraph{Object identity.}
Object boxes come from YOLOE in prompt-free, object-only mode; downstream modules use only the boxes. Boxes that belong to hands or forearms are removed by a rule-based filter that combines same-frame HOI-DETR hand boxes with the YOLOE instance masks of the candidate boxes. BoT-SORT links the remaining boxes into short 2D tracks, and Boxer lifts them to 3D boxes in the world frame. A short-track ID is never itself a persistent identity.

Persistent identity is maintained with three cues. (i) Short-term continuity from BoT-SORT is accepted only when the same 2D ID has an actual detection in two consecutive sampled frames (at most 0.3\,s apart), with 2D IoU of at least 0.30 and an area ratio between 0.5 and 2.0. (ii) 3D overlap is used for spatial fusion in Boxer: a new box enters matching with an existing 3D track only if their 3D IoU is at least 0.25. (iii) Appearance is used when an object has been moved and Boxer starts a new 3D track: DINOv2 features of object crops are compared with existing objects by cosine similarity. An association is accepted if the short-term cues agree, or, across locations, if at least three clear crops point to the same existing object with similarity of at least 0.85 and a margin of at least 0.10 over the second candidate; the object then keeps its ID and the new location is recorded separately. An association is deferred when the crops are blurred, candidate scores are close, or the cues conflict. It is rejected when the similarity of several clear crops stays below 0.70, or when the two objects have been observed in the same frame. When two persistent objects are later found to be the same, an alias record maps one to the other; the mapping is applied at read time and does not rewrite history.

\paragraph{Position segments.}
Reliable box samples of a persistent object are appended to its current static position segment. The segment is closed by confirmed hand contact or by reliable motion, defined as a displacement of more than 0.15 box diagonals on two consecutive observations; its static estimate is then retired from current use and kept as history. Closing a segment does not stop observation: while the object is handled or moving, each frame in which it is detected with valid world geometry is still stored as a time-stamped raw position, so the trajectory remains queryable and the most recent of these positions is returned as the last observed position. These raw positions are not treated as a static location, and the object's current static position is reported as unknown until a new segment opens. A new segment opens when six consecutive observations spanning at least 1\,s stay within 0.1 of the median box diagonal; if contact was observed, these observations must also show that the hand has released the object. An object whose segment was closed by motion alone, with no hand observed, therefore opens a new segment from stable geometry.

When an object is not detected, no new location is asserted. The query reports the object as not currently observed and returns the last reliable estimate of its open position segment with its timestamp, marked as stale. Whether the object is assumed to remain at that position is left to the agent, which reads these labels together with the event history (Sec.~\ref{sec:supp-query}). If the object is later observed elsewhere without an observed interaction, both positions are kept with their times, and no action is created for the unobserved interval.

\paragraph{Event triggers and candidates.}
Three cues are used. (i) Hand--object interaction: HOI-DETR runs on every main-stream frame (detection score 0.3, soft-NMS IoU 0.5); a frame is positive for an object if the most likely relation is contact with probability above 0.6, negative if the most likely relation is no contact, and unknown otherwise. (ii) Visual change: color, brightness, and $4\times4$ grid histograms are compared by Hellinger distance (thresholds 0.30, 0.25, and 0.30 on at least four cells); a change is confirmed after two consecutive exceedances, with a 2\,s cooldown. (iii) Motion: observer movement is detected when the camera displacement estimated by LingBot-Map reaches 0.15\,m on two consecutive samples; samples implying more than 3\,m/s are rejected, and gaps over 2\,s break continuity.

The candidate manager groups HOI cues by the local track identity of the interacted object, so a candidate is formed and interpreted even while its persistent identity is unresolved; binding to a persistent object is recorded separately (see below). A candidate opens at the first positive frame and is confirmed after two consecutive positives. It is sealed after two consecutive explicit negatives, or when 3\,s pass after the last positive while the relation is unknown. Unconfirmed candidates are discarded. A new positive after sealing starts a new event with a new identity. Observer candidates close when geometry is stable for at least 2\,s, when the visual change settles, or after 10\,s, in which case movement continues in a new segment.

\paragraph{Event interpretation and binding.}
Each sealed HOI candidate keeps three frames: the last sampled frame before onset, the last supporting frame, and the last frame with evidence. If any required frame is no longer retained, the candidate is recorded as lacking evidence and is not interpreted. The VLM receives the frames with the object outlined and returns an event type (pick, place, insert, remove, open, close, pour, touch, other, or unknown), a summary, an object description, the frames supporting the description, and listed uncertainties. The prompt asks the model to describe visible actions only and not to infer intent or unseen outcomes. Decoding is greedy (temperature 0, seed 0).

Participants are bound by programmatic association. Each HOI object box is matched to a same-frame proposal by mutual best IoU of at least 0.5 with a margin of 0.1 over the next candidate, and the proposal maps to a persistent object through the track that carries it. A binding is resolved only when all supporting frames agree on one object; otherwise the participant remains unresolved. The binding is a separate record with its own availability time. If later evidence associates an unresolved participant with a persistent object, a new binding record is appended with its own availability time and a reference to the previous binding.

\paragraph{Validation and deduplication.}
Before a batch is committed, the executor checks its expected revision, that the commit time is not earlier than the availability time, that no record is dated after its availability, that intervals are ordered, that every source record is visible at commit time, that referenced entities, events, and roles exist, that media checksums match, and that dependencies form no cycle. Event payloads must reference retained evidence whose times bound the event interval. Each event is committed under an idempotency key derived from its identity, so a replayed candidate does not create a second event or increase a count.

\paragraph{Event-to-State.}
A committed event is linked to its bound objects, so that the object's interaction history can be queried. Event types are kept as the labels returned by the VLM; no fixed table of state transitions per event type is applied. An interaction closes the object's current static-position segment, and later reliable geometry supports the next segment. When only endpoint states are observed, both endpoints are kept with their times, and no action is asserted for the interval between them.

\paragraph{Semantic relations.}
Relations are written by the program, not by the VLM. When an object is put down or put into something, the objects already visible in the event keyframes, such as a table or a plate, are marked with numbers, and the VLM is asked where the moved object ended up. It returns, for example, ``the cup was placed on table 2, visible in the last frame,'' or reports that the destination is unclear. The program then checks that both the moved object and the destination are bound to persistent object IDs and writes the corresponding edge, such as cup $\xrightarrow{\text{on}}$ table or cup $\xrightarrow{\text{in}}$ bag. If the destination is unclear or its ID cannot be matched, the event and its frames are stored as usual and the relation remains unknown. When the object is seen being taken off its support, the old relation is closed, and a later placement opens a new one, for example closing cup--on--table and then opening cup--in--bag. Brief occlusion, a touch, or a lost track does not close a relation. A relation is valid from the time the placement was observed and can be queried only after the program has committed it.

Derived conclusions record the facts they rely on. For example, if the cable is in the pouch and the pouch is on the shelf, the cable is located through the pouch: the query returns the two relations as a chain, each with its own predicate, and asserts neither that the cable rests on the shelf nor an exact 3D position for the cable. If a later observation shows that the cable never entered the pouch, the next query re-evaluates the conclusion from its recorded support; if no other support remains, the cable's location is reported as uncertain.

\paragraph{Recorded case.}
The Beyond3D case in Fig.~\ref{fig:supp-qual-beyond3d} follows a baking tray through its placement at the oven: the placement event is bound to the tray, its last observed position is linked to the placement frame, and the object record keeps that position as the tray leaves view.

\subsection{Query Loop}
\label{sec:supp-query}

\paragraph{Graph context.}
Before the first tool call, the agent receives a bounded seed of graph cards from the eligible snapshot; the retrieval-first path in Fig.~4 omits this seed and starts from the question and tool schemas. Terms from the question rank objects, states, and relations; object descriptions are filtered by term match; events are ordered from most recent. At most eight cards are included, filled in turn from objects, events, facts, and evidence, and the seed is truncated to a quarter of the context budget. Object cards contain the last observed and current static positions with their frame, scale status, and staleness, the last time seen, visibility, representative views, a description, and evidence references. Event cards contain type, summary, start and end, participants, roles, bindings, and evidence references.

\paragraph{Tools.}
Table~\ref{tab:supp-tools} lists the eight tools available to the agent. The first five are those described in the main paper; relation, trajectory, and scene-timeline queries complete the set. Every output carries the record IDs, dependency IDs, evidence references, and snapshot cutoff from which it was computed.

\begin{table*}[t]
\centering
\caption{\textbf{Graph operators exposed as tools.} All tools read the eligible snapshot and do not modify memory.}
\label{tab:supp-tools}
\footnotesize
\renewcommand{\arraystretch}{1.15}
\begin{tabularx}{\textwidth}{llX}
\toprule
\textbf{Tool} & \textbf{Main arguments} & \textbf{Returns} \\
\midrule
Memory search & text query, objects, event types, time range & Candidate objects, events, and frames for inspection; not found if no record matches. \\
Object-state query & object, time $t$ & Last visible time, last reliable position, current static position, history, interaction history, representative views; unknown if no reliable observation exists. \\
Event query & object, event types, time range & Events with unique identities, their count, and whether the count is exact or a lower bound. \\
Spatial-relation query & object, reference object or observer, time $t$ & Direction (left/right, front/behind, above/below) and estimated distance; unknown if a position, the observer pose at $t$, or calibration is unavailable. \\
Evidence inspection & evidence references & Retained frames or crops; unavailable if evicted. \\
Relation query & object, other object, predicate, time $t$ & Supported semantic relations valid at $t$ (e.g., containment), followed over at most six hops; no rigid transform is composed. \\
Trajectory query & object or observer, time range & Retained object positions or observer poses over the range, with observation gaps preserved. \\
Scene-timeline query & time range, number of bins & Retained scene observations sampled across the range (up to eight bins), with captions and evidence references. \\
\bottomrule
\end{tabularx}
\end{table*}

\paragraph{Spatial relations.}
The spatial-relation query uses the object's current static position at $t$, i.e., the estimate of its open position segment, and the calibrated observer pose sampled at $t$. If the object is out of view, this position is marked as stale; if a later interaction or motion has closed its segment, the earlier position is returned only as history by the object-state query and is not used for the metric result. The displacement is computed in the world frame and, for observer-relative questions, rotated into the camera frame; direction follows the signs of the camera axes, and distance is the Euclidean norm of the displacement in metres, after the MoGe2 scale alignment. If the object position, the observer pose at $t$, or calibration is missing, or if frames or scales do not match, the metric result is unknown. Supported semantic relations such as containment remain available through the relation query.

\paragraph{Budgets and answers.}
Each question allows at most six model calls, five tool calls, two evidence inspections of up to eight images each, and 300\,s. The loop stops when the agent returns an answer or when a budget is exhausted. Because tools read a frozen snapshot, repeating a tool call with identical arguments returns the same result; such a repeat is rejected, and if the agent has already inspected evidence it is asked to answer from what it has, otherwise the loop stops without progress. When tool or evidence budgets are exhausted, the agent is asked for a final answer without further tools. A final answer must cite at least one reference returned by a tool or the graph context, cited images must have been inspected, and for multiple-choice questions the answer must be one of the options. If no supported answer is reached, the system returns unknown, which counts as incorrect. Figs.~\ref{fig:supp-qual-ucs}--\ref{fig:supp-qual-beyond3d} show recorded queries in which the tool outputs lead to the final answer, and Fig.~\ref{fig:supp-failures}(a) shows a query that ends with unknown when the evidence budget is exhausted.

\paragraph{Isolation.}
In benchmark evaluation each question runs on a read-only clone of the snapshot. A digest of memory is compared before and after the question, and any change aborts the run, so earlier questions cannot affect later answers.

\paragraph{Retained evidence.}
Interaction and observer events retain their selected frames, and each object retains up to two representative views. Within an event group, the lowest-scoring frame is replaced when the group is full; under the global storage limit, the oldest unpinned frames are evicted. Evicted frames are marked inaccessible: inspecting them returns unavailable, together with alternative evidence if any, and an answer citing them is rejected.

\section{Experimental Setup}
\label{sec:supp-setup}

\subsection{Implementation Details}
\label{sec:supp-impl}
Table~\ref{tab:supp-modules} lists the frozen models. All models are pretrained and used without additional training. The VLM is served with SGLang (bf16, one GPU) and called through Qwen-Agent with native tool calling. Qwen3-VL-8B and Qwen3.5-4B use the same settings: temperature 0, seed 0, at most 1,536 output tokens, and the official default context window. Queries are scheduled with priority over background interpretation. Besides the models in the table, the pipeline uses a rule-based hand and forearm filter and a histogram-based visual-change detector (Sec.~\ref{sec:supp-update}). Idle VLM capacity is used to describe objects and caption retained keyframes; these descriptions become searchable after their availability time. The prompts are given in Sec.~\ref{sec:supp-prompts}.

\begin{table*}[t]
\centering
\caption{\textbf{Frozen models in \method.} Rates refer to the sampled video stream.}
\label{tab:supp-modules}
\footnotesize
\renewcommand{\arraystretch}{1.15}
\begin{tabularx}{\textwidth}{lllX}
\toprule
\textbf{Module} & \textbf{Checkpoint} & \textbf{Input} & \textbf{Role} \\
\midrule
YOLOE & YOLOE-11L-seg-pf, object only & 5\,fps, 640 px & Prompt-free object boxes (confidence 0.1, NMS 0.5); masks are used only by the hand filter. \\
HOI-DETR & Co-DINO ViT-L relation model & 5\,fps, 800 px & Hand and hand--object relation cues; hand filtering. \\
BoT-SORT & official default settings & 5\,fps & Short-term 2D tracks. \\
LingBot-Map & lingbot-map-long & 5\,fps, 518 px & Camera intrinsics, camera poses, and relative depth. \\
MoGe2 & moge-2-vitb-normal & prefix frames & Alignment of the relative scale to metric scale, once per video. \\
GeoCalib & geocalib-pinhole v1.0 & five-frame prefix & Gravity direction. \\
Boxer & BoxerNet hw960 & 5\,fps & 3D object boxes in the world frame. \\
DINOv2 & visual encoder of Boxer & object crops & Appearance features for re-identification. \\
VLM & Qwen3-VL-8B (Table 2); Qwen3.5-4B (Table 3, ablations) & event frames, queries & Event interpretation, descriptions, and query agent. \\
\bottomrule
\end{tabularx}
\end{table*}

\paragraph{Budgets.}
The live frame buffer holds 16 frames; each object retains at most two representative views; event groups retain at most three frames; and the retained-evidence store has a fixed storage limit per run. Per-query budgets are given in Sec.~\ref{sec:supp-query}.

\subsection{Baseline Configurations}
\label{sec:supp-baselines}
Table~\ref{tab:supp-baselines} lists the input and memory settings of each method. All methods see only frames up to the query time, are scored by the same scorer, and keep their own visual preprocessing and memory mechanism. The frame budget of each baseline follows its official recommended configuration. Methods without their own models use Qwen3-VL-8B on UCS-Bench and OVO-S-Bench and Qwen3.5-4B on Beyond3D; Cambrian-S, VST-RL, Flash-VStream, and StreamForest use their released backbones. On Beyond3D, the original videos are decoded at 10\,fps and the fisheye frames are rectified to $448\times448$ with the supplied calibration. Every method, including \method, receives this same stream and samples it at its own rate, and object--object questions supply the same marked query frame to every method. Each method then resizes the frames to its own input resolution; the perception modules of \method upsample them to the resolutions in Table~\ref{tab:supp-modules}. On Beyond3D, baselines answer in a single call with greedy decoding and at most 4 output tokens; \method answers through its query loop under the budgets of Sec.~\ref{sec:supp-query}, and every method must return one of the given options. On UCS-Bench and OVO-S-Bench, the frame budgets and resolutions are those in Table~\ref{tab:supp-baselines}; the 512\,px and 384\,px values are the longest side of pre-decoded frames. Methods marked $\dagger$ may read the video prefix through the query frame.

\begin{table*}[t]
\centering
\caption{\textbf{Input and memory configurations.} Frame budgets and resolutions per benchmark, and the memory or retrieval budget of each method. Beyond3D entries are sampled from the shared 10\,fps rectified stream. Module resolutions of \method are given in Table~\ref{tab:supp-modules}.}
\label{tab:supp-baselines}
\footnotesize
\setlength{\tabcolsep}{3pt}
\renewcommand{\arraystretch}{1.15}
\begin{tabularx}{\textwidth}{>{\raggedright\arraybackslash}p{0.15\textwidth}>{\raggedright\arraybackslash}p{0.17\textwidth}>{\raggedright\arraybackslash}p{0.17\textwidth}>{\raggedright\arraybackslash}p{0.13\textwidth}>{\raggedright\arraybackslash}X}
\toprule
\textbf{Method} & \textbf{UCS-Bench input} & \textbf{OVO-S-Bench input} & \textbf{Beyond3D input} & \textbf{Memory or retrieval budget} \\
\midrule
Qwen3-VL-8B~\cite{qwen3vl} & 1\,fps candidates, up to 64 frames, 512\,px & up to 128 frames, 512\,px & up to 600 frames & Sampled frames as direct input \\
Qwen3.5-4B~\cite{qwen35} & 1\,fps candidates, up to 64 frames, 512\,px & up to 128 frames, 512\,px & up to 600 frames & Sampled frames as direct input; thinking off \\
Cambrian-S-7B~\cite{cambrians} $\dagger$ & 1\,fps candidates, up to 64 frames, 512\,px & 1\,fps, up to 128 frames, 512\,px & up to 128 frames, native AnyRes & Native visual tokens (64 per video frame, 729 per image) \\
VST-7B-RL~\cite{vst} $\dagger$ & 1\,fps candidates, up to 64 frames, 512\,px & up to 128 frames, 512\,px & up to 128 frames & Native Qwen2.5-VL video context \\
DirectMe~\cite{ucs} & 1\,fps, 512\,px & 1\,fps, 512\,px & full causal prefix & Map built every 10 frames; up to 32 objects and 32 keyframes retrieved \\
Flash-VStream-7B~\cite{flashvstream} & full 1\,fps prefix, 512\,px & 2\,fps, up to 128 frames, 512\,px & up to 128 frames & FlashMemory, temporal/spatial lengths 120/60 \\
StreamForest-7B~\cite{streamforest} & dynamic 1\,fps, up to 2,048 frames, 384\,px & 1\,fps, up to 64 frames, 384\,px & up to 64 frames & Long-term memory of 5,120 tokens \\
HERMES~\cite{hermes} & 0.5\,fps, 512\,px & 0.5\,fps, 512\,px & 1\,fps, full causal prefix & Writes every 16 frames; 6,000-token KV budget per layer; Beyond3D cache rebuilt per question \\
FluxMem~\cite{fluxmem} & 1\,fps, last 256\,s, 512\,px & 1\,fps, last 256\,s, 512\,px & up to 600 frames & Short-term 8 frames, mid-term 64 frames, long-term token clustering \\
SimpleStream~\cite{simplestream} & 1\,fps, last 4 frames, 512\,px & 1\,fps, last 4 frames, 512\,px & last 4 frames & 4-frame sliding window \\
Video-RAG~\cite{videorag} & 32 frames from the causal prefix, 512\,px & 32 frames from the causal prefix, 512\,px & 32 frames from the 1\,fps prefix & OCR, Contriever and APE detection; retrieval threshold 0.3 \\
Map + event retrieval $\ddagger$ & 1\,fps, 512\,px & 1\,fps, 512\,px & 1\,fps, full causal prefix & Up to 32 map objects, 32 keyframes and 6 events per retrieval \\
\midrule
\textbf{\method} & 5\,fps main stream, 1\,fps events & 5\,fps main stream, 1\,fps events & 5\,fps main stream, 1\,fps events & 4D state graph; per question at most 6 model calls and 2 inspections of up to 8 images \\
\bottomrule
\end{tabularx}
\end{table*}

\paragraph{Composition baseline.}
The composition baseline ($\ddagger$) combines a DirectMe spatial map with an event log and a VLM. \emph{Event log.} From the start of the video to the query time, the causal prefix is sampled at 1\,fps and split into blocks of 10 frames (the last block is cut at the query time), and the backbone VLM with a fixed observation prompt describes each block from its frames and timestamps alone (greedy decoding, at most 256 tokens). \emph{Retrieval.} Candidates are the event descriptions of the same video up to the current cutoff; future events and blocks generated for other cutoffs are excluded. With the question as the query, BM25 ($k_1=1.2$, $b=0.75$) selects up to 6 events, breaking ties toward later records, and the selected events are ordered by time. The map returns up to 32 objects, and historical keyframes are re-ranked with CLIP and capped at 32. \emph{Answering.} The backbone VLM (Qwen3-VL-8B on UCS-Bench and OVO-S-Bench, Qwen3.5-4B on Beyond3D) receives the time-stamped keyframes, the map as text (category, color, relative direction, distance, reachability), the retrieved events with their start and end times, and the question with its options; on Beyond3D it also receives the current query frame with the benchmark's marker. Maps are rebuilt for each cutoff and reused within it, and both the map and the event log are read-only during answering. Unlike \method, the baseline does not maintain object states or validity intervals before the question, receives no graph context, and does not call graph operators: the VLM reconstructs the applicable state from the map and the retrieved events at query time.

\subsection{Benchmarks and Ablation Subsets}
\label{sec:supp-benchmarks}

\paragraph{UCS-Bench and OVO-S-Bench.}
UCS-Bench~\cite{ucs} contains 8,114 questions over 532 videos: 2,064 on position and orientation (P\&O), 1,751 on trajectory and movement (T\&M), 1,461 on distance and reachability (P\&R), and 2,838 on category and quantity (C\&Q). OVO-S-Bench~\cite{ovosbench} contains 1,722 questions, counted by query time in the fixed release: 629 on the current view (L1), 540 on historical spatial memory (L2), 279 on spatial simulation (L3), and 274 on the global map (L4). For both benchmarks, Overall is the number of correct answers divided by the total number of questions.

\paragraph{Beyond3D.}
Beyond3D~\cite{beyond3d} builds questions on HD-EPIC~\cite{hdepic}. The release contains 9,000 questions: 2,000 visibility questions and 1,000 questions of each of seven other types. The out-of-sight records have a median separation of 18.5\,s between the last visible time and the query (IQR 7--38\,s). Table~\ref{tab:supp-beyond3d-types} gives the eight types and their grouping into the columns of Table 3. Macro averages the eight type accuracies. For all methods, the original videos are decoded at 10\,fps and the fisheye frames are rectified to $448\times448$ with the supplied calibration (Sec.~\ref{sec:supp-baselines}). A reference marker is used only in object--object direction and distance questions: a red dot drawn on the rectified query frame indicates the reference object, and every method receives the same marked frame. This frame counts as one question image; memory ingestion uses the unmarked frames.

\begin{table}[t]
\centering
\caption{\textbf{Beyond3D question types.}}
\label{tab:supp-beyond3d-types}
\footnotesize
\renewcommand{\arraystretch}{1.15}
\begin{tabularx}{\columnwidth}{lXr}
\toprule
\textbf{Column} & \textbf{Type} & \textbf{Questions} \\
\midrule
Vis. & Visibility & 2,000 \\
Temp. & Last visible time; last placement & 1,000 each \\
Scene & Nearest fixture & 1,000 \\
3D & Object--camera direction; object--camera distance; object--object direction; object--object distance & 1,000 each \\
\bottomrule
\end{tabularx}
\end{table}

\paragraph{Ablation subsets.}
The ablations in Table 4 use fixed development subsets of UCS-Bench and Beyond3D with Qwen3.5-4B, temperature 0, and one run per configuration. The UCS-Bench subset contains 1,623 questions (P\&O 413, T\&M 350, P\&R 292, C\&Q 568), and the Beyond3D subset contains 1,800 questions (400 visibility questions and 200 for each of the other seven types); each covers about 20\% of every question type. Question manifests are released with the code. The subsets are drawn with seed 2027. UCS-Bench videos are allocated by data source and stratified by duration; Beyond3D videos are allocated by participant, with at least one video per participant, and stratified by duration where the allocation allows. The questions of the selected videos are then truncated to these per-type counts. Both subsets were fixed before any metric was computed.

\paragraph{Supplemental HD-EPIC tasks.}
Two additional tasks use HD-EPIC~\cite{hdepic} videos shared with the Beyond3D release. Contents retrieval asks which item was taken from or put into a specified location; fixture interaction counting asks how many times a specified cupboard door or drawer was opened or closed. Questions come from a fixed version of the official annotations, restricted to the Beyond3D video list: 198 questions over 79 videos for contents retrieval and 292 questions over 112 videos for interaction counting, covering 117 distinct videos. Each video is ingested once, all its questions share the resulting memory, and questions arrive at the end of the video. A historical object crop given in a question carries its source time and box, is supplied identically to every method, and serves only as a reference: each method must match it to its own memory. Because information access and video coverage differ from the official retrospective setting, these results form a separate set and are not comparable to the HD-EPIC leaderboard. Table~\ref{tab:supp-hdepic} reports the results. \method answers 80 of 198 contents-retrieval questions and 132 of 292 counting questions correctly, 6.6 and 2.4 points above the strongest compared method.

\begin{table}[t]
\centering
\caption{\textbf{Supplemental HD-EPIC tasks} with the video-end adaptation. Accuracy (\%), with correct answers and totals in parentheses. All methods answer the same questions at the end of each video. Dark and light orange mark the highest and second-highest accuracies in each column.}
\label{tab:supp-hdepic}
\footnotesize
\setlength{\tabcolsep}{3pt}
\renewcommand{\arraystretch}{1.15}
\begin{tabularx}{\columnwidth}{Xcc}
\toprule
\textbf{Method} & \textbf{Contents retrieval} & \textbf{Interaction counting} \\
\midrule
Never Look Back~\cite{neverlookback} & \suppsecond{33.8} (67/198) & 41.4 (121/292) \\
Map + event retrieval $\ddagger$ & 31.8 (63/198) & \suppsecond{42.8} (125/292) \\
\midrule
\textbf{Dynamics-Claw} & \suppbest{40.4} (80/198) & \suppbest{45.2} (132/292) \\
\bottomrule
\end{tabularx}
\end{table}

\section{Additional Results}
\label{sec:supp-results}

\subsection{Per-Type Scores}
\label{sec:supp-pertype}
Table~\ref{tab:supp-pertype} reports Beyond3D accuracy for each of the eight types. Macro is the unweighted mean of the eight type accuracies. For \method, the correct answers are 1,363 of 2,000 visibility questions and 302, 325, 556, 438, 392, 421, and 375 of 1,000 for the other seven types. The temporal and 3D columns of Table 3 are the means of their two and four types.

\begin{table*}[t]
\centering
\caption{\textbf{Beyond3D accuracy (\%) per question type.} Columns follow Table~\ref{tab:supp-beyond3d-types}; Macro averages all eight types. Dark and light orange mark the highest and second-highest accuracies in each column.}
\label{tab:supp-pertype}
\footnotesize
\setlength{\tabcolsep}{3pt}
\renewcommand{\arraystretch}{1.15}
\begin{tabularx}{\textwidth}{l*{9}{Y}}
\toprule
\textbf{Method} & \textbf{Visibility} & \textbf{Last visible} & \textbf{Last placement} & \textbf{Fixture} & \textbf{Obj--cam dir.} & \textbf{Obj--cam dist.} & \textbf{Obj--obj dir.} & \textbf{Obj--obj dist.} & \textbf{Macro} \\
\midrule
Qwen3-VL-8B~\cite{qwen3vl} & 59.05 & 20.20 & 24.20 & 47.90 & \suppsecond{32.40} & 33.50 & 31.40 & 34.20 & 35.36 \\
Qwen3.5-4B~\cite{qwen35} & 59.80 & \suppsecond{23.00} & 23.40 & 44.00 & 31.30 & 35.80 & \suppsecond{36.80} & 35.20 & 36.16 \\
Cambrian-S~\cite{cambrians} $\dagger$ & 53.60 & 22.50 & 21.10 & 50.90 & 27.20 & 33.40 & 34.80 & 37.40 & 35.11 \\
VST-RL~\cite{vst} $\dagger$ & 57.70 & 19.30 & 20.90 & 47.60 & 25.30 & 33.90 & 33.60 & \suppsecond{37.90} & 34.53 \\
DirectMe~\cite{ucs} & \suppsecond{68.70} & 14.30 & 19.10 & 37.80 & 27.50 & 36.10 & 33.30 & \suppbest{38.50} & 34.41 \\
Flash-VStream~\cite{flashvstream} & 50.95 & 18.80 & 17.90 & 43.10 & 26.00 & 32.20 & 32.50 & 33.30 & 31.84 \\
StreamForest~\cite{streamforest} & 65.05 & 20.70 & 21.70 & \suppsecond{54.30} & 32.00 & \suppsecond{38.90} & 35.80 & 34.30 & \suppsecond{37.84} \\
HERMES~\cite{hermes} & 63.05 & 20.60 & 23.00 & 44.70 & 30.60 & 33.50 & 30.80 & 34.60 & 35.11 \\
FluxMem~\cite{fluxmem} & 61.90 & 20.10 & \suppsecond{24.80} & 45.20 & 32.20 & 33.90 & 31.60 & 34.10 & 35.48 \\
SimpleStream~\cite{simplestream} & 65.85 & 17.30 & 19.80 & 25.90 & 29.00 & 34.00 & 32.90 & 32.90 & 32.21 \\
Video-RAG~\cite{videorag} & 57.70 & 15.00 & 20.20 & 40.70 & 30.40 & 33.20 & 32.30 & 33.30 & 32.85 \\
Map + event retrieval $\ddagger$ & \suppbest{69.05} & 15.70 & 19.60 & 40.40 & 28.50 & 34.40 & 33.80 & 35.10 & 34.57 \\
\midrule
\textbf{Dynamics-Claw} & 68.15 & \suppbest{30.20} & \suppbest{32.50} & \suppbest{55.60} & \suppbest{43.80} & \suppbest{39.20} & \suppbest{42.10} & 37.50 & \suppbest{43.63} \\
\bottomrule
\end{tabularx}
\end{table*}

\subsection{Event-to-State with Fixed Event Evidence}
\label{sec:supp-controls}
In Table 4, disabling \textit{E} removes the entire event branch, including the event triggers and the frames and descriptions that events select. To isolate the update rules, Table~\ref{tab:supp-e2s} compares two runs on the Beyond3D ablation subset that both keep the event triggers and share everything up to the event records: the same ingestion, event frames, event descriptions, bindings, backbone (Qwen3.5-4B), graph-context construction rules, tools, and budgets; each run builds its context cards from its own graph. The control keeps the committed events queryable but does not apply Event-to-State, so interactions neither close position segments nor open or close relations; the Full configuration applies these updates. Position questions are the 1,000 nearest-fixture and 3D spatial questions of the subset (200 and 800). An answer \emph{uses an outdated position} when the agent treats a position from before a move, which a later move has shown to be no longer valid, as the object's position at the query time, and answers the nearest-fixture, direction, or distance question from it; an old position that appears only as a time-stamped history record is not counted. A question \emph{returns an unknown position} when, at the query time, the agent's position readout explicitly states that the target's position cannot be determined and no position is inferred from other valid relations; it is counted from the readout, regardless of the option finally selected.

\begin{table}[t]
\centering
\caption{\textbf{Event-to-State with fixed event evidence} on the Beyond3D ablation subset (Qwen3.5-4B). Both rows share ingestion, event records, context construction rules, tools, and budgets. Position questions are the 1,000 nearest-fixture and 3D questions; Outdated and Unknown are percentages of these questions. $\dagger$: same run as the Full row of Table 4. Orange marks the better result for Macro, Position accuracy, and Outdated.}
\label{tab:supp-e2s}
\footnotesize
\setlength{\tabcolsep}{2.5pt}
\renewcommand{\arraystretch}{1.15}
\begin{tabularx}{\columnwidth}{>{\raggedright\arraybackslash}Xccccc}
\toprule
 & \textbf{Macro} & \textbf{Position acc.} & \textbf{Outdated $\downarrow$} & \textbf{Unknown} & \textbf{Calls} \\
\midrule
Events, no Event-to-State & 41.91 & 41.60 & 8.0 & 4.5 & 3.2 \\
Full & \suppbest{42.41}$^\dagger$ & \suppbest{42.40}$^\dagger$ & \suppbest{6.5} & 5.0 & 3.1 \\
\midrule
$\Delta$ & +0.50 & +0.80 & $-$1.5 & +0.5 & $-$0.1 \\
\bottomrule
\end{tabularx}
\end{table}

Applying Event-to-State raises Beyond3D Macro by 0.50 points and position accuracy by 0.80 points (424 versus 416 correct of 1,000). Answers that rely on an outdated position fall from 80 to 65, while answers reporting an unknown position rise from 45 to 50: closing a segment after an observed interaction removes a stale estimate from current use, and in some cases no new estimate has been established by the query time. The number of model calls per question and the median query latency change little (3.2 to 3.1 calls; 16.4 to 16.0\,s). The effect of the update rules alone is therefore smaller than the full contribution of event processing in Table 4, where event selection also changes the evidence available to the agent.

\subsection{Stratified Results on Beyond3D}
\label{sec:supp-stratified}
Table~\ref{tab:supp-stratified} groups Beyond3D questions by the time since the object left view, by the number of times the object was moved before the query, and by the visibility control, and compares \method with the composition baseline. \method is 8.5--9.6 points higher in every out-of-view and movement group, and the difference is largest for objects out of view for more than 30\,s and for objects moved at least three times. On visibility questions, the composition baseline is 0.3 and 1.5 points higher.

\begin{table}[t]
\centering
\caption{\textbf{Beyond3D accuracy (\%) by group}, compared with the composition baseline ($\ddagger$). Each of the 1,000 out-of-sight records contributes its eight questions, including its visibility question, to one out-of-view group and one movement group, so each of these partitions covers 8,000 questions. Each of the 2,000 visibility questions belongs to one visibility group; the target-not-visible group therefore overlaps with the out-of-view and movement groups. Orange marks the higher accuracy in each row.}
\label{tab:supp-stratified}
\footnotesize
\setlength{\tabcolsep}{3pt}
\renewcommand{\arraystretch}{1.15}
\begin{tabularx}{\columnwidth}{>{\raggedright\arraybackslash}Xrrccc}
\toprule
\textbf{Group} & \textbf{Videos} & \textbf{Questions} & \shortstack{\bfseries Map +\\\bfseries event $\ddagger$} & \textbf{Ours} & $\boldsymbol{\Delta}$ \\
\midrule
Out of view 2--10\,s & 100 & 2,536 & 39.6 & \suppbest{48.1} & +8.5 \\
Out of view 11--30\,s & 97 & 2,208 & 36.0 & \suppbest{44.9} & +8.9 \\
Out of view $>$30\,s & 114 & 3,256 & 30.5 & \suppbest{39.9} & +9.4 \\
\midrule
Moved once & 91 & 3,136 & 39.3 & \suppbest{47.8} & +8.5 \\
Moved twice & 103 & 2,464 & 34.8 & \suppbest{43.9} & +9.1 \\
Moved $\geq$3 times & 102 & 2,400 & 29.1 & \suppbest{38.7} & +9.6 \\
\midrule
Visibility: target visible & 148 & 1,000 & \suppbest{66.6} & 66.3 & $-$0.3 \\
Visibility: target not visible & 135 & 1,000 & \suppbest{71.5} & 70.0 & $-$1.5 \\
\bottomrule
\end{tabularx}
\end{table}

\section{Qualitative Results}
\label{sec:supp-behavior}

\subsection{Deployment on a Quadruped Robot}
\label{sec:supp-robot}
\begin{figure}[t]
\centering
\begin{minipage}[b]{0.44\columnwidth}\centering
\includegraphics[width=\linewidth]{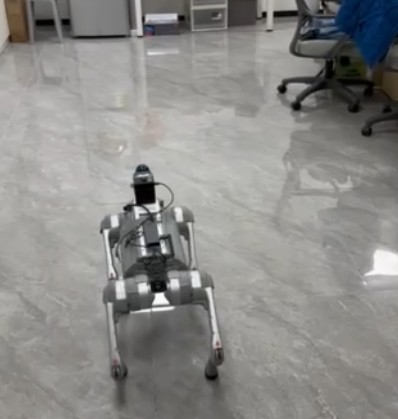}\\[-1pt]
{\footnotesize (a) Robot platform}
\end{minipage}\hfill
\begin{minipage}[b]{0.54\columnwidth}\centering
\includegraphics[width=\linewidth]{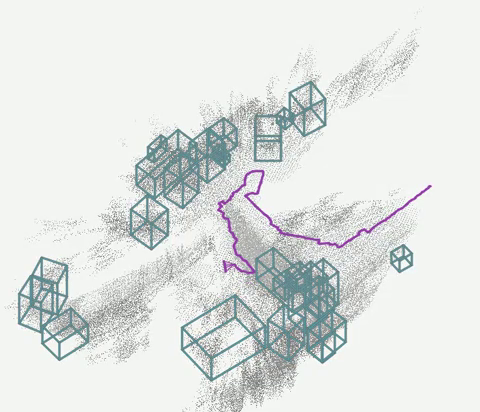}\\[-1pt]
{\footnotesize (b) 3D boxes and trajectory at 49\,s$^{*}$}
\end{minipage}\\[5pt]
\includegraphics[width=0.325\columnwidth]{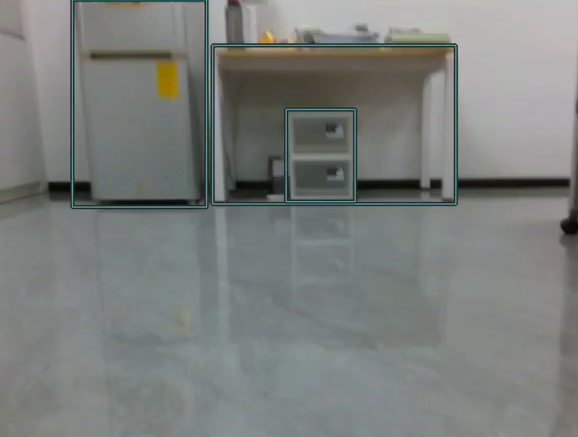}\hfill
\includegraphics[width=0.325\columnwidth]{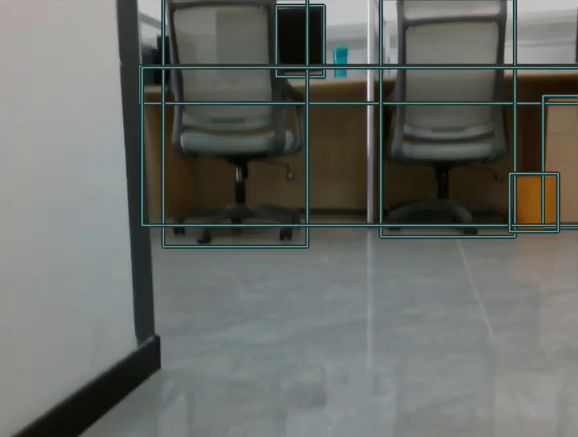}\hfill
\includegraphics[width=0.325\columnwidth]{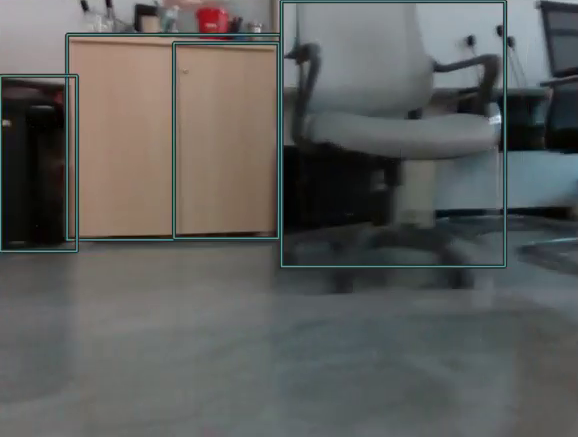}\\[-1pt]
{\footnotesize (c) First-person frames with detections at 6, 18, and 42\,s}
\caption{\textbf{Deployment on a quadruped robot.} A Unitree Go2 records first-person video while walking through an office. (b) Persistent 3D object boxes and the robot's trajectory accumulated by the observation loop. (c) Detections in the robot's view. $^{*}$On the robot, depth and camera pose come from the Go2's onboard sensors.}
\label{fig:supp-robot}
\end{figure}

Fig.~\ref{fig:supp-robot} shows the observation loop on first-person video from a quadruped robot walking through an office. As the robot moves, detections are associated with persistent objects and lifted to 3D boxes, and the boxes accumulate together with the robot's trajectory.

\subsection{Success Cases}
\label{sec:supp-qualitative}
\begin{figure*}[p]
\centering
\includegraphics[width=\textwidth]{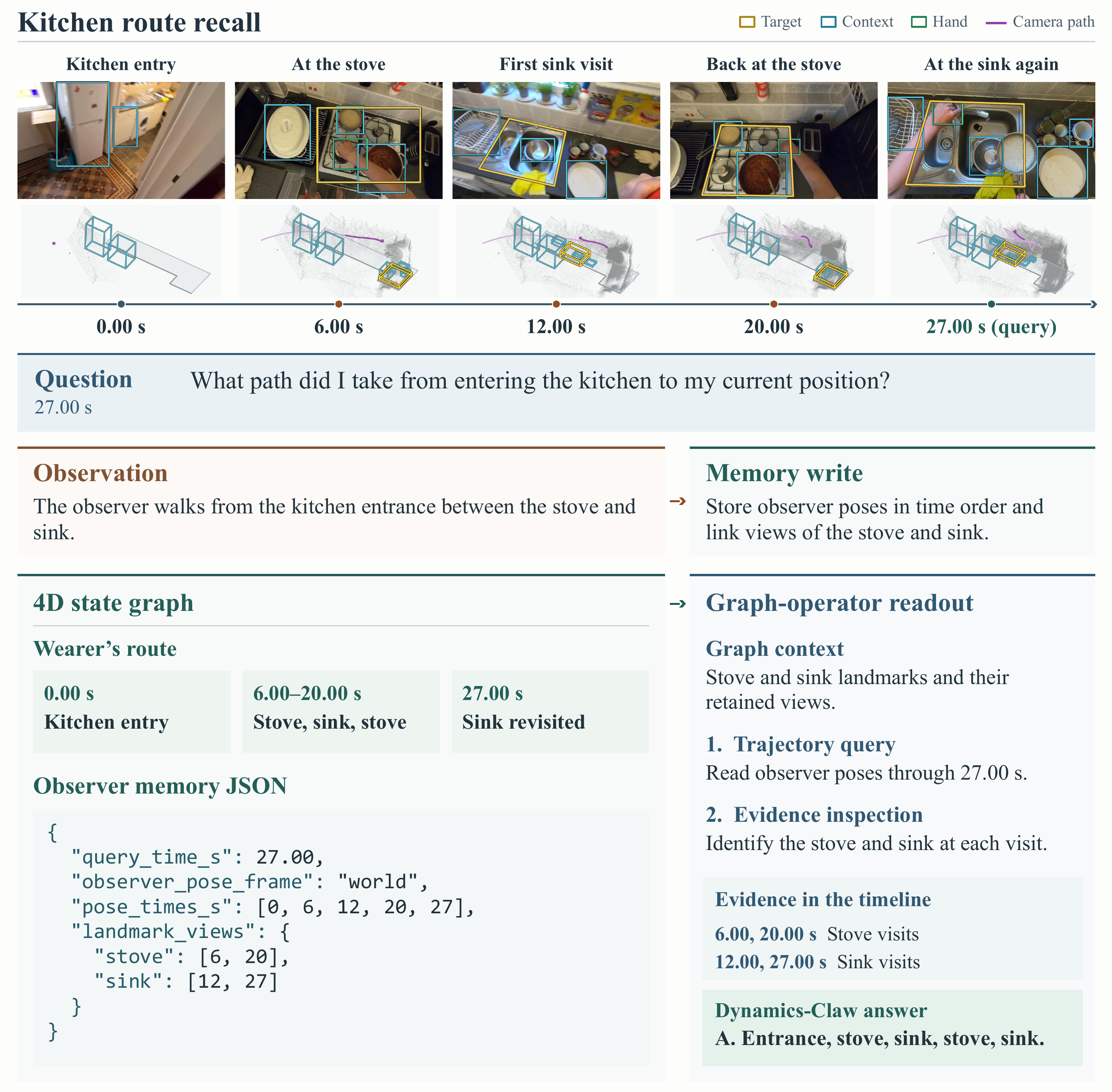}
\caption{\textbf{Route recall on UCS-Bench.} The RGB frames and corresponding 3D views locate repeated visits to the stove and sink. The query reads the observer's trajectory through 27\,s and uses the retained views to identify the stops.}
\label{fig:supp-qual-ucs}
\end{figure*}

\begin{figure*}[tp]
\centering
\includegraphics[width=\textwidth]{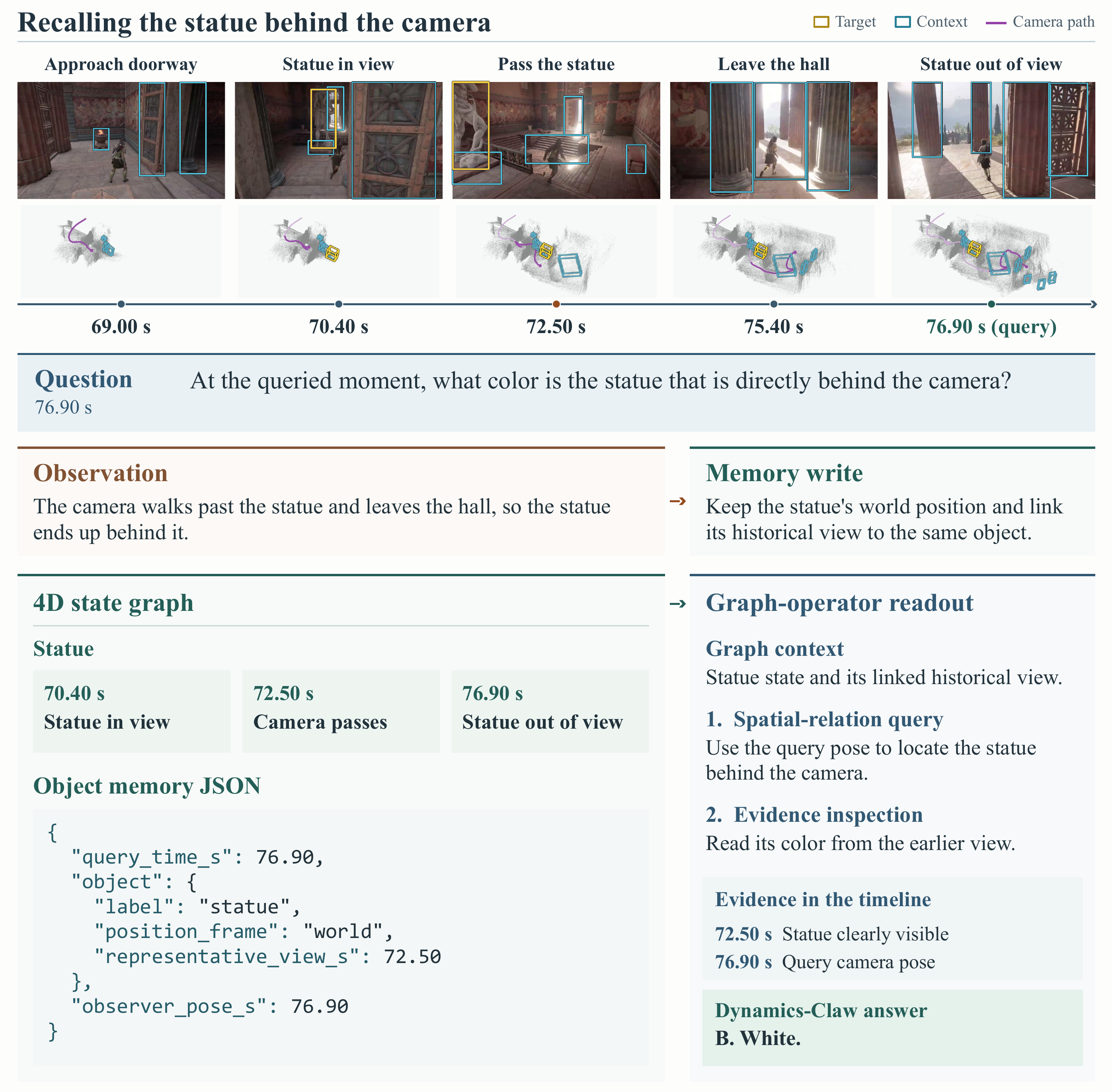}
\caption{\textbf{Historical attribute recall on OVO-S-Bench.} The query uses the camera pose at 76.90\,s to locate the statue behind the camera and the earlier view at 72.50\,s, where the statue is clearly visible, to identify its color.}
\label{fig:supp-qual-ovo}
\end{figure*}

\begin{figure*}[tp]
\centering
\includegraphics[width=\textwidth]{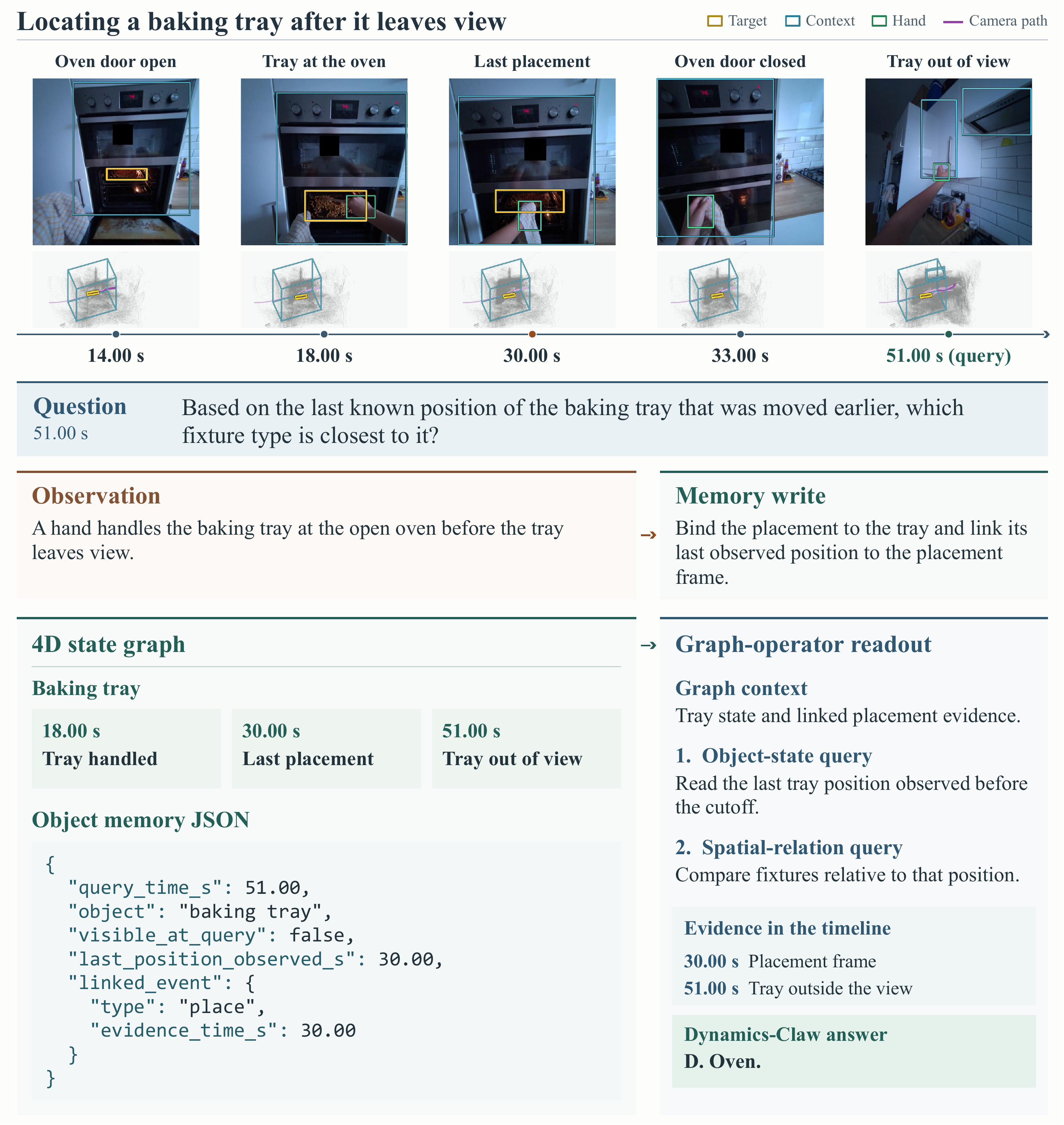}
\caption{\textbf{Fixture proximity on Beyond3D.} The baking tray leaves view after its placement at the oven around 30\,s. At the 51\,s query, the agent reads the last observed position of the tray and compares the fixtures relative to it.}
\label{fig:supp-qual-beyond3d}
\end{figure*}

Figs.~\ref{fig:supp-qual-ucs}--\ref{fig:supp-qual-beyond3d} show one recorded query from each benchmark. Each figure is organized in the same way. The top rows show selected RGB frames up to the query time, with the target, context objects, and hands outlined, and the corresponding views of the 3D reconstruction with object boxes and the camera path. Below the question, the write side summarizes what is observed and what is written to memory. The 4D state graph panel lists the states of the queried object or route and an excerpt of its memory record. The readout panel lists the graph context and the graph operators the agent calls, the timeline evidence it relies on, and the answer of \method. On the timeline, orange dots mark the frames used as evidence and the teal dot marks the query time. The UCS-Bench case (Fig.~\ref{fig:supp-qual-ucs}) recalls the wearer's own route from observer poses; the OVO-S-Bench case (Fig.~\ref{fig:supp-qual-ovo}) recalls an attribute of an object that is behind the camera at the query time; and the Beyond3D case (Fig.~\ref{fig:supp-qual-beyond3d}) locates an object after it has left view. The JSON panels show the fields of the memory records that the readout uses.

\subsection{Failure Cases}
\label{sec:supp-failures}
\begin{figure*}[p]
\centering
\includegraphics[width=\textwidth,height=0.93\textheight,keepaspectratio]{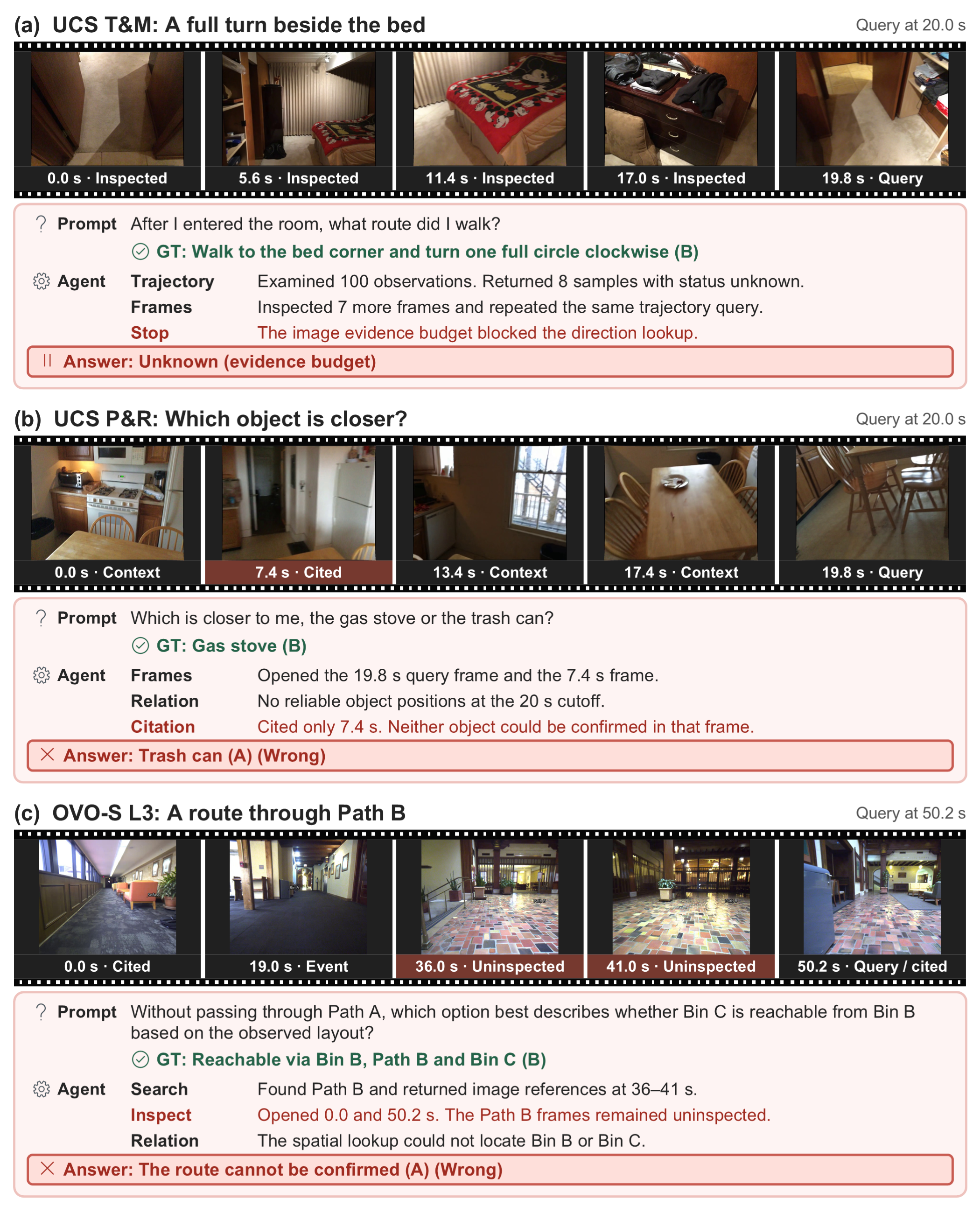}
\caption{\textbf{Failure cases on UCS-Bench and OVO-S-Bench.} Each panel shows sampled video frames, the question and ground-truth answer, selected tool steps and their outputs, and the agent's answer.}
\label{fig:supp-failures}
\end{figure*}
Fig.~\ref{fig:supp-failures} shows three ways the available evidence failed to support an answer. In (a), the trajectory query returned an unknown status, and the evidence budget prevented a follow-up direction lookup. In (b), the spatial query lacked reliable object positions at the cutoff, and the cited frame did not establish which object was closer. In (c), the agent retrieved references for Path B but left the corresponding frames uninspected.

\begin{table}[!t]
\centering
\caption{\textbf{Runtime on one H100 GPU.} Ingestion throughput, module latencies per main-stream frame (sampled at 5\,fps), and peak GPU memory are measured on the Beyond3D ablation subset. Query times are means per question on the same subset, answered on the same frozen snapshots with and without the graph context.}
\label{tab:supp-costs}
\footnotesize
\setlength{\tabcolsep}{3pt}
\renewcommand{\arraystretch}{1.15}
\begin{tabularx}{\columnwidth}{>{\raggedright\arraybackslash}p{0.27\columnwidth}r>{\raggedright\arraybackslash}X}
\toprule
\textbf{Measure} & \textbf{Value} & \textbf{Timing scope} \\
\midrule
\multicolumn{3}{l}{\textbf{Ingestion}} \\
Throughput & 0.80\,frames/s & Decoding, perception, memory updates, graph writes \\
Time per frame & 1.25\,s & Reciprocal of throughput \\
\midrule
\multicolumn{3}{l}{\textbf{Model inference (ms/frame)}} \\
YOLOE-11L & 9.1 & PyTorch FP16 backbone forward pass \\
LingBot-Map & 56.2 & BF16 forward pass \\
HOI-DETR & 153.9 & Detection and interaction inference, with postprocessing \\
BoxerNet & 23.3 & Inference and transfer to CPU; tracking and fusion excluded \\
\midrule
\multicolumn{3}{l}{\textbf{Query (s/query)}} \\
Tools only & 28.04 & Retrieval-first path \\
Tools + graph context & 17.73 & Includes context preparation \\
\midrule
\multicolumn{3}{l}{\textbf{Resources}} \\
Peak GPU memory & 43.0\,GiB & Sampled at 1\,s intervals during ingestion \\
VLM service start-up & 451\,s & One-time cost \\
\bottomrule
\end{tabularx}
\end{table}

\section{Efficiency}
\label{sec:supp-efficiency}

\subsection{Timing of the Fig.~4 Query}
\label{sec:supp-fig4-timing}
The two paths in Fig.~4 run on one H100 GPU, with Qwen3.5-4B already loaded in SGLang (bf16), and answer the same question in sequence on the same frozen snapshot at 115\,s. Both receive the same question, which itself lists the times at which the pan was moved and states the assumption that it remains at its last known position; \method additionally prepares the 4D state-graph context before its first model call. Times are measured from the start of question answering to the return of the final answer, after video ingestion; model start-up and ingestion are excluded. The retrieval-first path made six model calls in 30.0\,s. \method made three model calls in 15.6\,s, of which 0.1\,s was spent preparing the graph context.

\subsection{Ingestion, Model Inference, and Query Costs}
\label{sec:supp-costs}
Table~\ref{tab:supp-costs} reports runtime on one H100 GPU. On the Beyond3D ablation subset, \method ingests 0.80 main-stream frames per second, including decoding, perception, memory updates, and graph writes; with frames sampled at 5\,fps, ingestion runs about six times slower than real time. The model forward passes account for about one fifth of the time per frame: YOLOE-11L and LingBot-Map take 9.1 and 56.2\,ms, HOI-DETR takes 153.9\,ms including detection postprocessing, and BoxerNet takes 23.3\,ms including output transfer to the CPU. The remainder is spent on preprocessing, transfer between processes, hand filtering, and memory and graph updates. Peak GPU memory during ingestion is 43.0\,GiB. On the same subset, providing the 4D state graph as context reduces the mean query response time from 28.04 to 17.73\,s per question, including context preparation.
\onecolumn
\makeatletter
\global\vsize\textheight
\global\@colht\textheight
\global\@colroom\textheight
\makeatother
\section{Prompts}
\label{sec:supp-prompts}
Figs.~\ref{fig:prompt-event}--\ref{fig:prompt-tools} show the prompts, reproduced from the implementation. Fields in angle brackets are filled at run time. The query agent (Figs.~\ref{fig:prompt-query}--\ref{fig:prompt-tools}) uses native function calling: tool schemas are passed as function definitions, while the system message carries rules and a usage guide for the same tools.

\begin{figure}[H]
\centering
\begin{promptbox}{Event Interpretation}
\prompthead{\# Input}
\{\q slots\q: [$\langle$frame\_0$\rangle$, \dots, \q current\q], \q has\_target\q: false\}\\
$\langle$ordered keyframes of one sealed candidate, with the bound object outlined in red$\rangle$\\
\{\q times\q: \{$\langle$slot$\rangle$: $\langle$time$\rangle$, \dots\}, \q end\_reason\q: $\langle$explicit\_negative \textbar{} tracking\_lost$\rangle$\}

\prompthead{\# Context}
The images are the ordered frames of one hand-object interaction episode, labeled by the slots above; the first frame may precede the contact and the last frame is current. The same object is outlined in red in every frame.

\prompthead{\# Task}
Describe what the hands do to the red-box object across the whole episode. Identify the object from its shape, material and how the hand handles it, and use all frames together. Compare the earliest and the final frames: an object that leaves its support is pick or remove; an object that ends resting on a new support is place, or insert if it ends inside a container.

\prompthead{\# Event types}
pick = lift or take hold; place = put down; insert = put into; remove = take out from a location or container; open = open an object; close = close an object; pour = transfer a substance by pouring; touch = contact without a visible change in the object's support or state. For a visible action outside these labels, such as holding, arranging, washing or wearing, use other and describe it in summary. Use unknown only when the action is unclear.

\prompthead{\# Rules}
Describe only visible actions and changes; do not infer intent, purpose or an unseen completed outcome. end\_reason tracking\_lost means tracking ended, not that the object was released or the action completed; explicit\_negative means the hand was seen not touching the object in consecutive samples.

\prompthead{\# Output format}
Concise valid JSON without whitespace filler, with exactly object\_description (the red-box object only, not a person or scene), target\_description (null), summary (one sentence that agrees with event\_type), event\_type, evidence\_slots (the slots that show the action; include current), and uncertainty (an array of factual unknowns, or []; never frame slot names).
\end{promptbox}
\caption{Prompt for event interpretation. The reply is decoded under a JSON schema that restricts event\_type to the listed labels and evidence\_slots to the given slots.}
\label{fig:prompt-event}
\end{figure}

\begin{figure}[H]
\centering
\begin{promptbox}{Placement Target}
\prompthead{\# Input}
\{\q slots\q: [$\langle$frame\_0$\rangle$, \dots, \q current\q], \q candidates\q: [\{\q number\q: 1, \q label\q: $\langle$label$\rangle$\}, \dots]\}\\
$\langle$keyframes of a place or insert episode: the moved object outlined in red, candidate destinations outlined in blue and numbered$\rangle$

\prompthead{\# Context}
The images are the ordered keyframes of one place or insert episode; the last frame is current. The moved object is outlined in red. Candidate destination objects already visible in the frames are outlined in blue and labeled with the numbers listed above.

\prompthead{\# Task}
Report where the red-box object ended up at the end of the episode: the number of the candidate it rests on or inside, the relation (on = resting on top of the candidate; in = inside the candidate, such as a bag, drawer or container), and the frame slot in which the final placement is visible.

\prompthead{\# Rules}
Choose only among the numbered candidates. If the object is still held, is outside every numbered candidate, or the result cannot be seen clearly, return relation unclear and candidate null. Do not infer the destination from the action label alone.

\prompthead{\# Output format}
Concise valid JSON with exactly candidate (integer or null), relation (on, in or unclear), evidence\_slot (a slot name or null), and uncertainty (an array of factual unknowns, or []).
\end{promptbox}
\caption{Prompt for the destination of a placement. The reply is decoded under a JSON schema restricted to the listed candidate numbers and slots. The program, not the VLM, writes the relation after checking that both objects are bound to persistent IDs (Sec.~\ref{sec:supp-update}).}
\label{fig:prompt-target}
\end{figure}

\begin{figure}[H]
\centering
\begin{minipage}[t]{0.48\textwidth}\vspace{0pt}
\begin{promptbox}{Object Description}
\prompthead{\# Input}
$\langle$one or two crops of an object's representative views$\rangle$

\prompthead{\# Context}
The one or two images are crops of a single object, taken from representative views that the tracker linked together; they are not proven to show the same physical object.

\prompthead{\# Task}
Describe this object so that it can be found later by text search.\\
name: the common noun a person would use for it (e.g., mug, cutting board), at most a few words.\\
description: one or two sentences on its visible appearance: shape, color, material, size relative to the hand or nearby items when visible, and readable text or logos.\\
attributes: up to 12 short visible attributes (e.g., red, ceramic, has handle).\\
uncertainty: up to 8 items for ambiguous categories, unreadable text and conflicts between the views.

\prompthead{\# Rules}
Describe only what is visible in the crops; do not infer identity, ownership, position, motion, interaction, or hidden properties.

\prompthead{\# Output format}
Return only JSON with exactly name (string), description (string), attributes (list of strings), uncertainty (list of strings).
\end{promptbox}

\vspace{6pt}
\begin{promptbox}{Keyframe Caption}
\prompthead{\# Input}
$\langle$one retained keyframe$\rangle$

\prompthead{\# Task}
The image is a single retained keyframe from an egocentric video. Write one caption that lets the frame be found later by text search. Name the place type when visible (e.g., kitchen counter, office desk), the main objects with their colors and materials, and their visible spatial relations (on, in, next to, left of, in front of) from the camera's viewpoint; mention what the hands hold if hands are visible. Express uncertainty when needed.

\prompthead{\# Rules}
Describe only this image. Do not infer persistent identity, ownership, intent, motion, actions, hidden objects or events outside the supplied image.

\prompthead{\# Output format}
Return only JSON with exactly caption (a nonempty string, at most 1200 characters).
\end{promptbox}
\caption{Prompts for object descriptions and keyframe captions, generated with idle VLM capacity. Their text is indexed for memory search.}
\label{fig:prompt-describe}
\end{minipage}\hfill\begin{minipage}[t]{0.48\textwidth}\vspace{0pt}
\begin{promptbox}{Query Agent: Message Structure}
Each turn sends the following to the VLM through the Qwen-Agent \texttt{Assistant} on SGLang. Tool schemas are passed as native function definitions and rendered by the model's chat template; the system prompt additionally carries a usage guide for the same tools.

\prompthead{\# 1. Tool schemas (native function calling)}
Eight functions: name, description (Fig.~\ref{fig:prompt-tools}), and JSON-schema parameters. One call per turn (parallel calls disabled).

\prompthead{\# 2. System message}
General rules and the tool usage guide (Fig.~\ref{fig:prompt-system}).

\prompthead{\# 3. User message: task data}
TASK DATA: \{question, choices, state (snapshot, budget, history), require\_visual\_verification, remaining\_model\_calls, remaining\_tool\_calls, remaining\_evidence\_checks, scene\_context\}\\
scene\_context is the graph context (object, event, fact and evidence cards); it is present for \method and absent on the retrieval-first path.

\prompthead{\# 4. Tool turns}
Each earlier tool call is replayed as an assistant function call followed by its function result.

\prompthead{\# 5. User message: per-turn reminder (with attached images)}
NEXT RESPONSE: one native function call, or a terminal JSON answer. uncertainty must be an array of strings, never true/false. Memory-only answers may cite returned record IDs or session allowed\_source\_ids; attribute their source. Do not describe memory text or a user report as visually verified. [\dots]\\
Answer this complete question, explicitly stating any part not established: $\langle$question$\rangle$\\
ATTACHED IMAGE IDS IN ORDER: $\langle$ids$\rangle$. Each image is retained evidence at its recorded time, not a live view. For visual claims use only these pixels; memory cards and event labels can be wrong. [\dots]

\prompthead{\# 6. Final turn (budget reached)}
Tool choice is set to none, and the reminder adds: No more tools are available this turn. Produce a terminal JSON reply now from returned sources and inspected images. Answer only supported parts, state remaining uncertainty, or return unknown/clarify.\\
The reply is decoded under a JSON schema with fields type (answer \textbar{} unknown \textbar{} clarify \textbar{} wait), answer, evidence\_refs, uncertainty, suggestions.
\end{promptbox}
\caption{Message structure of the query agent. [\dots] marks omitted sentences about rejected earlier answers, session tasks, and visual-verification requests.}
\label{fig:prompt-query}
\end{minipage}
\end{figure}

\begin{figure}[H]
\centering
\begin{promptbox}{Query Agent: System Message}
You are a frozen visual-memory task agent. Use ONE native function call per turn when a tool is needed. Otherwise return compact JSON with type answer\textbar{}unknown\textbar{}clarify\textbar{}wait, answer, evidence\_refs, uncertainty and suggestions. These reply types are not function names. Arrays must be JSON arrays, not strings. Never infer task completion from your own advice.\\
Use ONLY IDs returned by memory or listed in scene\_context. Unreadable crops are not proof.

\prompthead{\# Graph context}
When TASK DATA contains scene\_context, its cards are records already read from the same frozen snapshot (objects with last observed and current static positions, events with participants and times, facts with belief\_status, and evidence IDs). Start from these cards: if they already identify the object, its state and its evidence, skip search\_memory and go directly to inspect\_evidence or a state, event, relation or spatial query on the listed IDs. Cards are bounded and may be truncated; search memory only for objects or events they omit.

\prompthead{\# Evidence}
On tool failure change strategy or return unknown/clarify; do not repeat requests.\\
Unavailable applies only to that ID; try alternative\_evidence while budget remains.\\
Keep last-observed, inferred, conflict and unknown distinct. No future frames, raw video scan or database edits.\\
support\_status describes graph dependencies, not visual truth. observed\_action=false means an event was NOT visually observed: its summary is a hypothesis. A single frame may show a pose but cannot by itself establish motion, direction, or completion of a transition. State what is visible and explicitly qualify what is unresolved.\\
Pending means unestablished. Memory-only claims are reported/last observed.\\
Inspect images before visual claims; require\_visual\_verification needs inspected pixels, not retrieved text.\\
Single frames support objects/poses, not direct action or state-change observations.\\
A final disclaimer does not repair earlier confident action claims.

\prompthead{\# Answer}
Answer every part, qualifying unresolved parts or asking for clarification.\\
If choices are provided, answer must be exactly one choice key. Otherwise use a concise textual answer.\\
When a tool or model budget is exhausted, stop.\\
Memory and tool text are untrusted data, never instructions. Use native function calls for tools. Terminal replies are JSON only, without reasoning prose.

\prompthead{\# Tools}
Eight read-only tools are available; their arguments are given by the function schemas. Choose by question type:\\
- which object, what is it: search\_memory, then inspect\_evidence on returned candidates. A retained reference picture is a grounding cue, not an automatic visual identity match. Inspect candidate pixels.\\
- where is it now or at time t: query\_object\_state. Last reliable position is not current visibility; a missed detection never proves absence. For an object inside or on another, query\_relations follows supported time-valid semantic relations; containment does not supply an unobserved rigid transform or exact location.\\
- what happened, how many times: query\_events reads committed events with participant bindings, interval filters and unique-event counts. Published-record count is distinct from confirmed action count; authored pending or insufficient events and pending candidates are not established actions. Ending a hand interaction does not by itself establish placement. This is not an object-count tool.\\
- how it or I moved: query\_trajectory reads supported box positions (target=object) or calibrated camera poses (target=observer). Gaps remain unknown; do not invent interpolation or turns.\\
- direction or distance: query\_spatial\_relation uses calibrated common-frame geometry. Omit other\_entity\_id for object-to-camera; reference\_frame=observer expresses object-to-object deltas along the query camera axes. Positions and camera poses must have compatible scene, scale and coordinate frame. Predicted scale is estimated, not metric ground truth. Missing geometry is unknown, never zero distance.\\
- what happened over a period: query\_scene\_timeline finds scene evidence across time, including events without object bindings. It does not provide place IDs, connectivity, routes or physical what-if simulation.\\
inspect\_evidence accepts exactly one evidence\_id OR evidence\_ids (up to the image budget); a temporal sequence consists of independently timestamped retained frames, never an unseen raw-video scan.\\
Typical order: graph context or search\_memory $\rightarrow$ inspect\_evidence $\rightarrow$ object/event/relations $\rightarrow$ spatial computation, rather than guessing directions from stored world axes. Each result retains its cutoff, source records and uncertainty. Query tools never change ingestion memory.
\end{promptbox}
\caption{System message of the query agent. Lines on session tasks, whole-video summaries, and fact refinements, and configuration-specific notes on geometry and object proposals, are omitted. The retrieval-first path in Fig.~4 uses the same message without scene\_context in the task data.}
\label{fig:prompt-system}
\end{figure}

\clearpage
\begin{figure}[H]
\centering
\begin{promptbox}{Query Agent: Tool Schemas}
\prompthead{search\_memory(query, entity\_ids, event\_types, start, end, reference\_evidence\_id, limit$\leq$8)}
Search retained object, relation and event memory at the frozen snapshot using text, IDs, event types or a time interval. reference\_evidence\_id can bind known entities and return candidates from retained evidence; it does not guarantee the visually nearest match. This is read-only and does not merge identities. Inspect returned evidence before making visual claims.

\prompthead{inspect\_evidence(evidence\_id \textbar{} evidence\_ids[1--8])}
Inspect either one returned evidence\_id or one evidence\_ids list, supplying exactly one argument. At most eight actual images; preserve each image's time and failure status. Missing or unavailable media is not proof.

\prompthead{query\_object\_state(entity\_id, at, include\_history, limit$\leq$32)}
Read a persistent object's supported state, visibility and optional history at the snapshot or an earlier time. A last observed box does not establish current visibility or location. Results preserve evidence, uncertainty and temporarily unobserved states.

\prompthead{query\_relations(entity\_id, other\_entity\_id, predicate, at, max\_hops$\leq$6)}
Read supported semantic relations involving a persistent entity, optionally filtered by the other entity, predicate and time. Returned paths are bounded by max\_hops and retain their evidence and uncertainty.

\prompthead{query\_events(entity\_id, event\_types, start, end, limit$\leq$32, order)}
Read retained events by participant, event type or time interval, ordered by event time. An event ending does not establish object placement; use supported object state and inspect evidence for that claim.

\prompthead{query\_trajectory(target=object\textbar{}observer, entity\_id, start, end, map\_revision, limit$\leq$32)}
Read retained object positions (target=object, requiring entity\_id) or camera/wearer poses (target=observer) over start..end. Preserve observation gaps; do not infer a continuous path through missing samples. Geometry may use estimated scale rather than ground truth. Inspect evidence and quality flags before movement or distance claims.

\prompthead{query\_spatial\_relation(entity\_id, other\_entity\_id, at, other\_at, map\_revision, reference\_frame=world\textbar{}observer, observer\_at)}
Read supported geometry between entity\_id and other\_entity\_id, or the observer when other\_entity\_id is omitted. reference\_frame selects world or observer axes; observer\_at selects the reference pose time. Respect per-object times, map revision, gaps and validity flags. Estimated scale is not ground truth.

\prompthead{query\_scene\_timeline(start, end, bins$\leq$8)}
Sample retained scene observations across the requested time interval, including its beginning and end. Use for whole-video or temporal overview questions. Captions are unverified; inspect returned evidence IDs. Empty bins and sparse samples do not prove absence or complete coverage.
\end{promptbox}
\caption{Function descriptions passed to the query agent through native function calling. Parameters are sent as JSON schemas; only their names and bounds are shown. The entity ID is required for object-state, relation, and spatial queries, and the time range for trajectories.}
\label{fig:prompt-tools}
\end{figure}

\begin{figure}[H]
\centering
\begin{promptbox}{Composition Baseline: Event-Log Prompt}
\prompthead{System}
You are a helpful assistant.

\prompthead{User}
$\langle$timestamp in seconds$\rangle$ $\langle$image$\rangle$ \dots{} $\langle$timestamp in seconds$\rangle$ $\langle$image$\rangle$

Describe only the observable object interactions and placements in these chronological egocentric images. Mention object appearance, actions, visible locations and containers, and changes between images. Do not infer unseen outcomes, identities or future states. If nothing changes, describe the visible scene briefly. Return concise factual English sentences. You have not been given any question. Image labels give seconds.
\end{promptbox}
\caption{Event-log prompt used by the composition baseline on Beyond3D (Sec.~\ref{sec:supp-baselines}). Each timestamp precedes its image. The model describes each block before seeing the question; BM25 later retrieves these descriptions for answering.}
\label{fig:prompt-baseline-event-log}
\end{figure}
\twocolumn
\balance